\documentclass[10pt, a4paper, logo]{googledeepmind} %
\usepackage{times}

\usepackage{hyperref}
\usepackage[leftcaption]{sidecap} 

\usepackage{amsmath,amsfonts,bm}
\usepackage{natbib}
\usepackage{fancyhdr}

\def\eqref#1{equation~\ref{#1}}

\def\1{\bm{1}}

\DeclareMathAlphabet{\mathsfit}{\encodingdefault}{\sfdefault}{m}{sl}
\SetMathAlphabet{\mathsfit}{bold}{\encodingdefault}{\sfdefault}{bx}{n}

\usepackage{hyperref}
\usepackage{url}
\usepackage{booktabs}
\usepackage{tabularray} 
\usepackage{graphicx} 
\usepackage{subcaption}
\usepackage{amssymb}
\usepackage[most]{tcolorbox}
\usepackage[table]{xcolor}
\tcbuselibrary{breakable} 
\usepackage{amsmath} 
\usepackage{siunitx}
\usepackage{colortbl}

\usepackage{microtype}
\usepackage{color}
\usepackage{wrapfig}
\usepackage{natbib}
\usepackage{colortbl}
\usepackage{microtype}
\usepackage{graphicx}
\usepackage{booktabs} 
\usepackage{array}
\usepackage{textcomp}
\usepackage{stfloats}
\usepackage{float}
\usepackage{verbatim}
\usepackage[ruled, linesnumbered, lined]{algorithm2e}
\usepackage{multirow}
\usepackage{enumitem}
\usepackage[utf8]{inputenc}
\usepackage{listings}
\usepackage{amssymb} 
\usepackage{caption}
\usepackage{threeparttable}
\usepackage{tabularx}
\usepackage{array}      
\usepackage{makecell}   
\usepackage[utf8]{inputenc}
\usepackage{enumitem}
\usepackage[most]{tcolorbox}
\usepackage{fontawesome5}
\usepackage{geometry}

\newcolumntype{C}{>{\centering\arraybackslash}X}

\definecolor{BestColor}{HTML}{C8E6C9}  
\definecolor{SecondBestColor}{HTML}{FFF9C4} 

\usepackage{amsmath,amsfonts}

\definecolor{ggg}{RGB}{26,179,0}
\definecolor{rrr}{RGB}{179,0,0}
\definecolor{oodc}{RGB}{31,73,121}
\definecolor{idc}{RGB}{68,142,68}

\definecolor{mygray}{gray}{0.9}

\def\Bias#1#2{\bm{b}}

\newtcolorbox{examplebox}[2][]{ 
    breakable, 
    enhanced, 
    colback=white, 
    colframe=cyan, 
    coltitle=white, 
    fonttitle=\bfseries, 
    title=#2, 
    overlay middle={\draw[cyan, line width=1pt](frame.south west)--(frame.south east);}, 
    overlay last={\draw[cyan, line width=1pt](frame.south west)--(frame.south east);}, 
    #1 
}

\usepackage[T1]{fontenc}
\usepackage{booktabs}      
\usepackage{graphicx}      
\usepackage[table]{xcolor} 
\usepackage{siunitx}       
\usepackage{etoolbox}      
\usepackage[normalem]{ulem}     

\definecolor{impcolor}{HTML}{2E8B57} 

\newcommand{\improvementstyle}[1]{$^{\textcolor{impcolor}{\tiny #1}}$}

\newcommand{\scoreimp}[2]{%
  \textbf{#1}%
  \ifstrequal{#2}{+0.0}{}{%
    \ifstrequal{#2}{0.0}{}{%
      \makebox[0pt][l]{\improvementstyle{#2}}%
    }%
  }%
}

\definecolor{headergray}{RGB}{245, 245, 245}
\definecolor{bordergray}{RGB}{220, 220, 220}
\definecolor{highlightgreen}{RGB}{144, 238, 144} 

\tcbuselibrary{listings, skins}

\tcbset{
    inputformatbox/.style={
        enhanced,
        listing only, %
        colback=white,
        colframe=bordergray,
        arc=5pt, 
        boxrule=1pt,
        top=1mm,
        bottom=1mm,
        left=1mm,
        right=1mm,
        title={#1},
        colbacktitle=headergray,
        coltitle=black,
        fonttitle=\large\bfseries\sffamily, 
        attach boxed title to top left={yshift=-3mm, xshift=3mm},
        boxed title style={
            sharp corners, 
            boxrule=0pt, 
            colframe=headergray
        },
        before title={\small\faCaretDown\quad},
        listing options={
            basicstyle=\ttfamily\small,
            breaklines=true,
            numbers=left,
            numberstyle=\tiny\color{gray},
            numbersep=8pt,
            xleftmargin=15pt,
            escapeinside={(*}{*)},
            showstringspaces=false
        }
    }
}

\title{QuarkMedSearch: A Long-Horizon Deep Search Agent for Exploring Medical Intelligence}

\author[1]{Zhichao Lin\textsuperscript{*\ddag}}
\author[1,2]{Zhichao Liang\textsuperscript{*\dag}}
\author[1]{Gaoqiang Liu}
\author[1]{Meng Xu}
\author[1]{Baoyu Xiang}
\author[1]{Shuxin Zhao}
\author[1]{Yao Wu}
\author[1]{Jian Xu}
\author[1]{Guanjun Jiang}

\affil[1]{Qwen Applications Business Group, Alibaba}
\affil[2]{Shanghai Jiao Tong University}

\begin{abstract}
As agentic foundation models continue to evolve, how to further improve their performance in vertical domains has become an important challenge. To this end, building upon Tongyi DeepResearch, a powerful agentic foundation model, we focus on the Chinese medical deep search scenario and propose QuarkMedSearch, systematically exploring a full-pipeline approach spanning medical multi-hop data construction, training strategies, and evaluation benchmarks to further push and assess its performance upper bound in vertical domains. Specifically, for data synthesis, to address the scarcity of deep search training data in the medical domain, we combine a large-scale medical knowledge graph with real-time online exploration to construct long-horizon medical deep search training data; for post-training, we adopt a two-stage SFT and RL training strategy that progressively enhances the model's planning, tool invocation, and reflection capabilities required for deep search, while maintaining search efficiency; for evaluation, we collaborate with medical experts to construct the QuarkMedSearch Benchmark through rigorous manual verification. Experimental results demonstrate that QuarkMedSearch achieves state-of-the-art performance among open-source models of comparable scale on the QuarkMedSearch Benchmark, while also maintaining strong competitiveness on general benchmarks.
\end{abstract}

\begin{document}
\maketitle

\section{Introduction}
With the rapid advancement of large language models (LLMs), the field is shifting away from monolithic general-purpose systems toward models optimized for agentic use. A new generation of frontier models—including GPT-5~\citep{singh2025openai}, Claude Opus 4.6~\citep{anthropic_claude_opus_4_6_2026}, Gemini 3.1 Pro~\citep{gemini3pro2025}, GLM-5~\citep{team_glm-5_2026}, and Kimi-K2.5~\citep{team_kimi_2026}—places long-horizon planning and tool orchestration at the center of model design. In parallel, the open-source community is rapidly narrowing the gap with commercial systems. Models such as MiroThinker~\citep{team_mirothinker_2025}, Tongyi DeepResearch~\citep{team_tongyi_2025}, and RedSearcher~\citep{chu_redsearcher_2026} have achieved state-of-the-art open-source results on agentic benchmarks including GAIA~\citep{mialon_gaia_2023}, BrowseComp~\citep{wei_browsecomp_2025}, and Humanity's Last Exam~\citep{phan2025humanity}. Against this backdrop, deep search has emerged as a core paradigm~\citep{openai2025o3deepresearch} for solving complex, knowledge-intensive tasks through autonomous planning, iterative retrieval, multi-step reasoning, and cross-document synthesis in real-world Web environments.

Medical science stands as one of the most compelling testbeds for AGI, demanding complex multi-step reasoning, authoritative knowledge synthesis, and high-stakes decision-making\citep{xu2026quarkmedicalalignmentholistic,wu2026quarkmedbenchrealworldscenariodriven}. Research on medical deep search, however, remains at an early stage. Existing training data is still dominated by general-domain tasks, leaving a severe shortage of high-quality long-horizon data for Chinese medical scenarios. As a result, current medical deep search models~\citep{yu_medresearcher-r1_2025, lu_med-r3_2026} struggle on BrowseComp-level tasks that require long-horizon planning and multi-round iterative retrieval. Their limitations stem from short interaction horizons, narrow data sources (e.g., English-only literature databases such as PubMed), and reliance on offline retrieval sources that do not reflect real Web environments. The problem is compounded by the fact that frontier models already internalize substantial medical knowledge during pretraining, allowing them to answer many questions directly from parametric memory rather than through retrieval. This weakens the training signal for deep retrieval and long-horizon reasoning. The medical domain further raises the bar through its specialized knowledge, the difficulty of cross-document synthesis, strict source-authority requirements, and the need for interpretable reasoning chains. Together, these factors expose a clear gap between current general-purpose agents and the demands of real medical use cases.

Evaluation remains similarly underdeveloped. Although benchmarks such as MedBrowseComp~\citep{chen_medbrowsecomp_2025} and MedXpertQA~\citep{zuo2025medxpertqa} have appeared in recent years, they share two common limitations. First, many are constrained in scenario design: their sources are often restricted to specific government websites or fixed retrieval channels and therefore do not capture the openness and complexity of real-world medical information seeking. Second, they emphasize internalized knowledge and short-horizon reasoning more than core agentic abilities such as multi-round tool use, dynamic retrieval-strategy adjustment, and long-horizon execution. As a result, the field still lacks a benchmark that comprehensively measures long-horizon agentic capability in open medical environments.

Beyond data and evaluation, another open problem is how to transfer agentic capability effectively into specialized domains. As foundation models become stronger agents, unlocking their performance ceiling in professional settings has become increasingly important. Medical scenarios are especially demanding: they require greater reasoning rigor, higher factual accuracy, more authoritative sources, and more reliable post-training and evaluation protocols than general-purpose tasks.

To address these challenges, we present \textbf{QuarkMedSearch}, a training framework for long-horizon deep search agents in the Chinese medical domain that covers the full pipeline from data synthesis and post-training to benchmark construction. Our main contributions are as follows:

\begin{itemize}[leftmargin=*,align=left]
\vspace{-2mm}
\item \textbf{Four-Phase Progressive Medical Deep Search Data Synthesis.}
We propose a four-stage pipeline for constructing long-horizon medical reasoning data. First, we mine rare medical entities from a large-scale medical knowledge graph grounded in real-world Web data and build a high-quality seed question set via subgraph sampling. Second and third, we perform multi-hop factual expansion and critical-entity obfuscation under an agentic paradigm, using online search and self-verification to deepen reasoning while suppressing shortcuts through parametric memory. Fourth, we apply rigorous multi-stage verification to filter for answer clarity and uniqueness, discarding samples with ambiguous or non-unique answers. The resulting corpus provides reliable long-horizon reasoning chains for training.

\item \textbf{SFT and RLVR Combined Post-Training Recipe for Long-Horizon Medical Agents.}
We propose a two-stage post-training recipe that combines supervised fine-tuning with reinforcement learning from verifiable rewards. SFT progresses from short to long trajectories to establish stable long-horizon search behavior, while RLVR further improves performance under a strict reward design that mitigates reward hacking. Throughout training, we mix general-domain and medical data to preserve broad capabilities. The resulting model consistently improves on the QuarkMedSearch Benchmark while remaining competitive on general deep search benchmarks.

\item \textbf{QuarkMedSearch Benchmark.}
We construct the QuarkMedSearch Benchmark with medical experts as a rigorous testbed for Chinese medical long-horizon deep search. It contains 140 human-verified questions spanning six major categories, with strict guarantees of answer uniqueness and reasoning depth.

\item \textbf{Transfer of Agentic Capabilities to Business Scenarios.}
We further validate that capabilities learned from short-answer medical deep search transfer to medical long-answer generation. In particular, long-horizon planning, iterative retrieval, and multi-step reasoning generalize effectively to real business scenarios and yield clear performance gains.

\end{itemize}

\section{Preliminary}

\paragraph{Problem Formulation.}
 We formulate the deep search task as a sequential decision-making process in which an agent operates within a real-world Web environment. Unlike single-turn question answering, deep search queries $q$ are knowledge-intensive and require multi-step reasoning—a single retrieval call is insufficient to yield a direct answer. The agent must therefore progressively accumulate evidence across multiple interaction rounds. At each step, it plans the next retrieval direction based on currently known information, invokes tools to obtain new evidence fragments, and dynamically updates its reasoning path. This cycle repeats until all reasoning constraints implicit in $q$ are satisfied and the target answer is determined. This stands in fundamental contrast to traditional Retrieval-Augmented Generation (RAG), which performs a single retrieval and directly feeds the results into the model for answer generation.

\paragraph{Core Variable Definitions.}
We adopt the standard ReAct framework~\citep{yao2023react} as the agent's underlying architecture, organizing the interaction process as an alternating sequence of (thought, action, observation) triples. We define the following core variables for a complete deep search session:
\begin{itemize}[leftmargin=*,align=left]
    \item \textbf{Query ($q$):} The information need posed by the user. In our setting, $q$ typically takes the form of a complex reasoning problem with long inference chains, strong constraints, and cross-entity dependencies.

    \item \textbf{Reasoning ($\tau_t$):} The agent's reasoning process prior to executing an action at step $t$, summarizing and planning based on currently collected evidence, already-satisfied constraints, and reasoning directions yet to be verified, without producing any external tool calls.

    \item \textbf{Action ($a_t$):} The tool invocation instruction issued by the agent at step $t$. The action space consists of three types of operations: \texttt{search}, \texttt{visit}, and \texttt{answer}, used respectively to retrieve result summaries, extract detailed content from a specified page, and directly produce the final answer when accumulated evidence sufficiently supports the reasoning conclusion.

    \item \textbf{Observation ($o_t$):} The environmental feedback returned by the tool after the agent executes action $a_t$, whose specific form depends on the type of tool invoked.

    \item \textbf{Answer ($y$):} The final response generated at the end of the interaction, which must be grounded in the collected evidence and satisfy all reasoning constraints implicit in query $q$.
\end{itemize}

\paragraph{Complete Interaction Process.}
Let $\mathcal{H}_{t-1} = \langle q, (\tau_0, a_0, o_0), \ldots, (\tau_{t-1}, a_{t-1}, o_{t-1}) \rangle$ denote the complete interaction history up to step $t$. At each step, the agent generates the next internal reasoning step $\tau_t$ and tool-call action $a_t$ conditioned on the current history:
\begin{equation}
    (\tau_t, a_t) \sim \pi(\cdot \mid \mathcal{H}_{t-1})
\end{equation}
The action $a_t$ is then executed by the tool executor, returning an observation $o_t = \mathcal{F}(a_t)$. The history is subsequently extended to $\mathcal{H}_t = \mathcal{H}_{t-1} \cup \{(\tau_t, a_t, o_t)\}$, which serves as the context for the next step. The complete reasoning trajectory is recorded as:
\begin{equation}
    \mathcal{H}_T = \langle q,\ (\tau_0, a_0, o_0),\ (\tau_1, a_1, o_1),\ \ldots,\ (\tau_T, a_T, o_T),\ y \rangle
\end{equation}
After $T$ steps of interaction, the agent generates the final answer $y = g(q, \mathcal{H}_T)$ based on the accumulated trajectory.

\section{Medical Deep Search Task Synthesis}
\label{sec:Medical Deep Search Task Synthesis}
\subsection{Overview}
\begin{figure}[H]
    \centering
    \includegraphics[width=\linewidth]{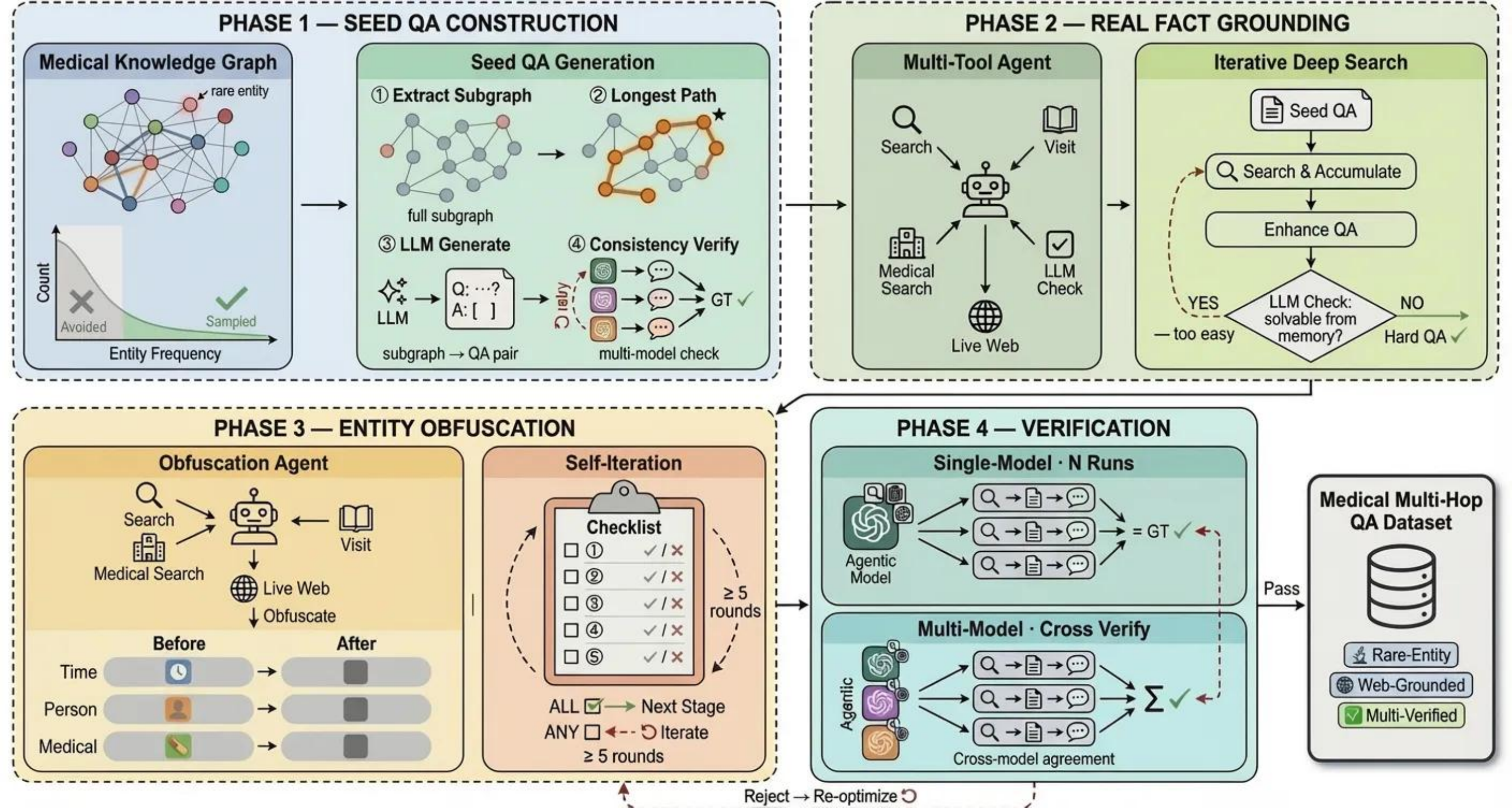}
    \caption{Medical Deep Search Task Synthesis Pipeline.}
    \label{fig:pipeline}
\end{figure}
Training an agent with medical deep search capabilities places stringent requirements on training data: it must feature multi-hop reasoning chains, complex question designs, and sufficient medical knowledge density, with key facts at each reasoning hop necessarily derived from external retrieval rather than parametric memory. Existing data, however, falls short along two dimensions: medical deep search datasets involve too few tool-call rounds, while general-purpose long-horizon deep search datasets contain only a negligible proportion of medical samples with insufficient domain coverage. These limitations jointly indicate that a dedicated data synthesis pipeline tailored to medical deep search is urgently needed.

To address this gap, we build a scalable and controllable pipeline for synthesizing long-horizon medical training data. As illustrated in Figure~\ref{fig:pipeline}, the pipeline has four modules: (1) \textbf{Seed Question Construction}, which uses long-tail medical entities and structured path sampling over an in-house medical knowledge graph to produce seed QA pairs with multi-hop reasoning skeletons; (2) \textbf{Multi-Hop Real-Fact Introduction}, which grounds these skeletons in real online evidence and forces each hop to depend on retrieval rather than internalized knowledge; (3) \textbf{Entity Obfuscation}, which systematically raises difficulty and retrieval necessity through critical-entity rewriting and iterative checklist-based refinement; and (4) \textbf{Uniqueness and Correctness Guarantee}, which audits each sample through multi-rollout and cross-model verification to ensure a unique, unambiguous answer.

\subsection{Phase 1: Knowledge Graph-Based Seed Question Construction}

\subsubsection{Long-Tail Entity Sampling}
\label{sec:Long-Tail Entity Sampling}

Unlike general-purpose domains, medical knowledge is characterized by high rigor and specialization: concepts are precisely defined, entity relationships are complex, and causal chains exhibit strict logical dependencies. These properties impose higher requirements on knowledge sources for seed question construction. Directly using raw web content as the basis risks introducing noise, entity ambiguity, and factual conflicts, which can distort question descriptions and blur answer boundaries, thereby undermining training data quality at the source. In contrast, triples in a structured medical knowledge graph are systematically curated, with entities and relations possessing clear semantic boundaries, providing a more reliable and semantically coherent knowledge foundation for seed question construction.

To this end, we construct a large-scale medical knowledge graph dedicated to medical reasoning tasks, integrating real-world internet medical resources with millions of professional medical book corpora to balance knowledge timeliness and professional depth. The graph covers a wide range of entity types—including diseases, drugs, gene targets, and anatomical structures—forming a multi-granularity medical knowledge network. However, LLMs exhibit a natural internalization bias toward high-frequency medical concepts during pretraining—common diseases, mainstream drugs, and standard treatment protocols appear frequently in training corpora, allowing models to answer directly from parametric memory without actually triggering retrieval or reasoning. Training samples centered on such high-frequency entities therefore fail to effectively elicit multi-hop retrieval and cross-source reasoning capabilities. To mitigate this bias, following MedResearcher-R1~\citep{yu_medresearcher-r1_2025}, we stratify all entities by frequency based on their in- and out-degree in the graph, and prioritize sampling from the long-tail region. These long-tail entities cover knowledge-sparse domains such as rare diseases, niche drug targets, and non-mainstream treatment protocols, making it difficult for the model to rely solely on internalized knowledge, thus ensuring that training signals more effectively act on the model's retrieval and reasoning capabilities. 

\subsubsection{Subgraph Extraction and Question Construction}

Starting from the long-tail entity identified in Section~\ref{sec:Long-Tail Entity Sampling} as the root node, we expand outward hop by hop along semantic relations—such as \textit{symptom findings}, \textit{associated findings}, \textit{complications}, \textit{contraindications}, and \textit{dietary recommendations}—to extract a local subgraph. For path selection, following MedResearcher-R1~\citep{yu_medresearcher-r1_2025}, we prioritize the longest semantic paths from the root node to maximize reasoning hops and knowledge span. Empirically, too few hops yield overly simple questions that frontier models can answer from parametric memory, while too many hops cause semantic divergence and undermine logical coherence and answer uniqueness. Balancing these considerations, we set the expansion depth to 4--6 hops.

Once the subgraph is determined, we use the terminal entity as the answer and the multi-hop reasoning conditions encoded in the subgraph as the question description, leveraging an LLM to transform the subgraph into natural language QA pairs. To ensure factual correctness, we introduce multiple frontier closed-source models to perform consistency evaluation on the ground truth. A sample passes verification only when multiple models converge on the same answer that is consistent with the graph facts; otherwise, the error reasons are fed back to the LLM for regeneration until consistency is achieved.

\subsection{Phase 2: Online Environment-Based Deep Search QA Expansion}

On the initial seed QA set, we conducted systematic difficulty evaluation experiments and found that frontier models can still leverage their powerful parametric memory to resolve most reasoning constraints—either by directly answering intermediate nodes through internalized knowledge, or by pinpointing the answer with only minimal retrieval, rather than undergoing the complete multi-hop reasoning process. The root cause lies in a distributional shift between seed questions generated from closed graphs and the real internet environment: question semantics are overly clean and standardized, lacking the noise and ambiguity inherent in real retrieval scenarios, while the semantic relations captured by graph triples are relatively direct, allowing models to complete cross-hop inference through knowledge analogy without actual retrieval. This results in a severe discrepancy between the apparent hop count of questions and the effective retrieval depth they actually trigger.

\begin{figure}[h]
    \centering
    \includegraphics[width=\linewidth]{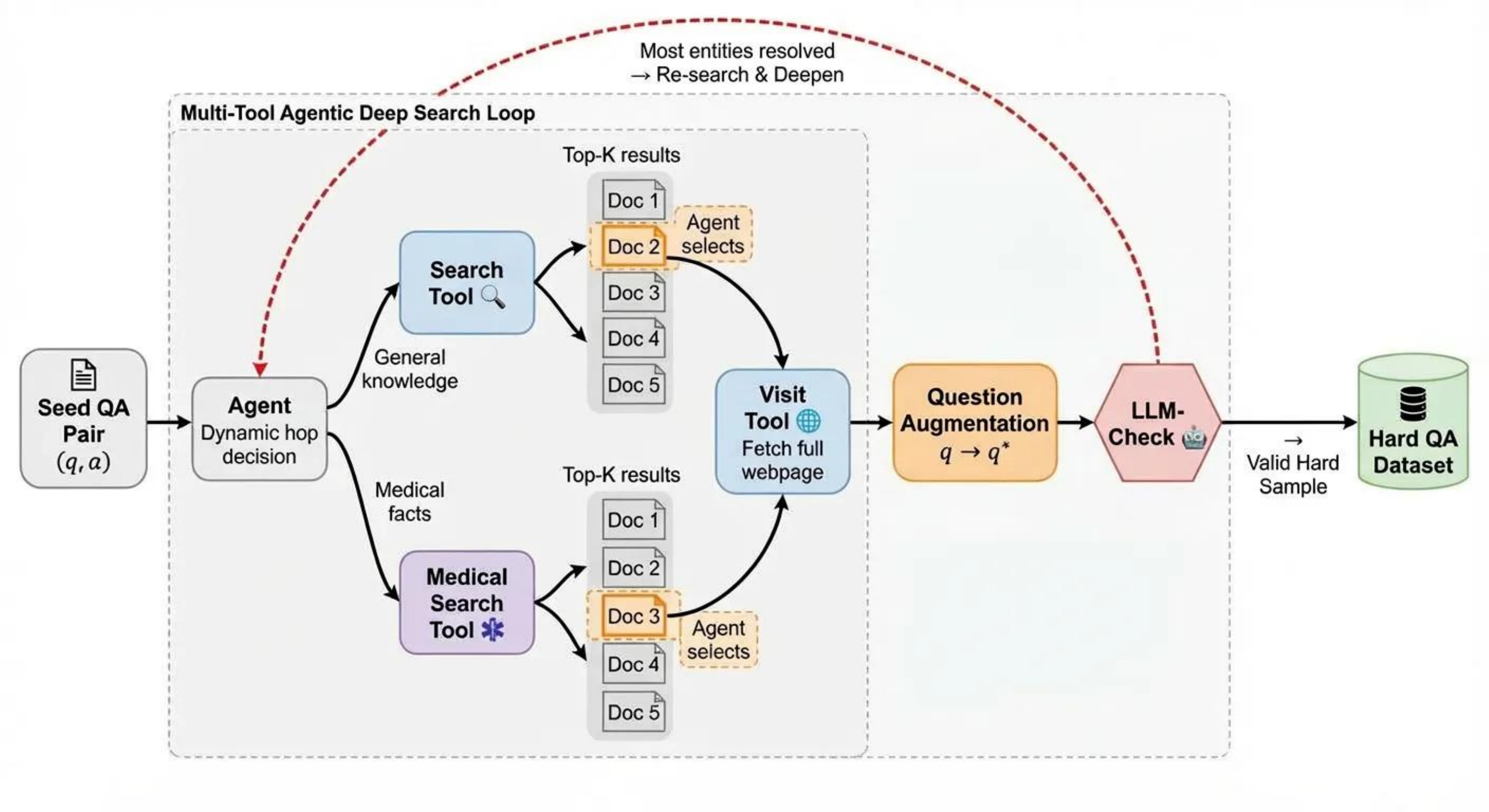}
    \caption{Long-Horizon Deep Search QA Augmentation in Online Environments.}
    \label{fig:phase2}
\end{figure}

To address this, building on the WebExplorer framework\citep{liu_webexplorer_2025}, we extend it by designing a multi-hop real-fact introduction module, illustrated in Figure~\ref{fig:phase2}. This module equips the model with four tools—Search, Visit, Medical Professional Search, and LLM Check (detailed in Section~\ref{sec:Multi-Tool Agentic Execution Framework}). Taking the seed QA pairs as input, the model advances hop by hop, invoking appropriate tools to introduce real online facts. Each time the model produces a new candidate question, LLM Check is used to verify whether a frontier model can still answer it from parametric memory. If so, the question is sent back for further deepening; otherwise, it is saved as a valid hard sample once the number of reasoning hops exceeds 10. Through this closed-loop mechanism, each reasoning hop undergoes retrieval-necessity verification, effectively reducing the likelihood that the model can complete the reasoning process solely through parametric memory.

\subsubsection{Multi-Tool Agentic Execution Framework}
\label{sec:Multi-Tool Agentic Execution Framework}
To support long-horizon deep search in real-world internet environments, we equip the execution model with four complementary tools:
\begin{itemize}[leftmargin=*,align=left]
    \item \textbf{Search.} Retrieves relevant documents from the real-world Web, returning top-$K$ candidate results as raw information sources.

    \item \textbf{Visit.} Fetches the full-text content of a candidate webpage, which is subsequently summarized by a dedicated summary model to extract information relevant to the current query.

    \item \textbf{Medical Professional Search.} A search tool specifically optimized for medical domain retrieval, with particular emphasis on source authority and information timeliness. 

    \item \textbf{LLM Check.} The best-performing model in the Qwen series is invoked to generate a direct response to the current query without any tool calls, so that the medical knowledge distributions across models remain broadly aligned from pretraining. The resulting reasoning trace and final answer are then passed to a judge model, which determines whether the key entities are directly resolved in the \texttt{think} component and correctly produced in the \texttt{answer} component. The resulting judgments are returned as rewards to guide subsequent action selection.

\end{itemize}

\subsubsection{Long-Horizon Deep Search Process Simulation}
As illustrated in Figure~\ref{fig:pipeline}, starting from an initial seed QA pair, the execution model operates as an agent that dynamically determines the next tool invocation based on its current reasoning state. The agent first searches for relevant facts and selectively reads those of interest to obtain more comprehensive information, where the medical professional search tool is prioritized for medical domain knowledge while the general search tool is invoked for background or general-domain queries. Once the agent judges that sufficient facts have been gathered, it constructs a new question and invokes the LLM check tool to assess the retrieval necessity of each key entity contained in the question. If the question fails to reach an effective challenge threshold, the agent analyzes the weak points and adjusts its search direction and enhancement strategy accordingly; otherwise, the question is admitted as a valid sample. Every multi-hop QA sample generated is anchored to verifiable real-world facts, with each reasoning hop having undergone retrieval necessity verification, effectively reducing the likelihood of reasoning solely through parametric memory and laying a solid foundation for subsequent training data quality.

\subsection{Phase 3: Entity Obfuscation}
\label{sec:obfuscation}

After multi-hop fact introduction, questions may still be answerable through straightforward search without requiring extensive multi-step retrieval. To address this, we apply systematic entity obfuscation to further raise question difficulty, achieving a level of challenge comparable to BrowseComp~\citep{wei_browsecomp_2025}. Since accurate substitution requires thorough understanding of each entity's attributes, the execution model proactively invokes search tools to collect background knowledge before performing any substitution. Different entity types are handled with differentiated strategies: temporal entities are converted to descriptive time ranges; location entities are replaced with functional descriptions; person entities are replaced with occupational or relational descriptions; medical terms are rewritten as symptom manifestations or pathological descriptions; and numerical indicators are replaced with relative or interval-based expressions. However, since obfuscation involves multiple competing quality objectives that are difficult to satisfy in a single pass, we further introduce a multi-round self-iterative checklist optimization mechanism to systematically review and refine the results.

\paragraph{Multi-Round Self-Iterative Checklist Optimization.}
After each round of obfuscation generation, the model conducts a systematic review along four dimensions:
\begin{itemize}[leftmargin=*,align=left]
    \item \textbf{Replacement Naturalness:} Whether the obfuscation substitution is natural and accurate, effectively concealing key information while maintaining semantic coherence.
    \item \textbf{Coverage Completeness:} Whether any unprocessed explicit entity cues remain in the question; if omissions are found, corresponding substitutions are added and verification is re-triggered.
    \item \textbf{Difficulty Effectiveness:} Whether the obfuscated question remains too simple, i.e., whether there is a risk that the model can directly derive the answer from internalized knowledge alone.
    \item \textbf{Answer Uniqueness:} Whether the current obfuscation strategy has inadvertently compromised answer uniqueness, ensuring that the obfuscated question still corresponds to a single unambiguous ground-truth answer.
\end{itemize}
A minimum of five iteration rounds is enforced to avoid local optima, and a question proceeds to the next phase only when all checklist items pass and this minimum is reached, ensuring consistent obfuscation quality across all generated samples.

\subsection{Phase 4: Uniqueness and Correctness Guarantee}
\label{sec:verification}

Following entity obfuscation, the compound operations of multi-hop fact introduction and repeated paraphrase rewriting may introduce semantic ambiguity, answer drift, or logical inconsistencies that individual stage-level quality checks cannot fully eliminate. Given the dataset scale, manual verification is prohibitively expensive, so we employ multiple state-of-the-art agentic models as automated verifiers for systematic quality auditing. However, single-model single-pass verification has inherent limitations: stochastic inference may yield inconsistent results across reasoning trajectories, and model-specific knowledge biases may produce systematic misjudgments. To address this, we conduct verification along three complementary dimensions:

\begin{itemize}[leftmargin=*,align=left]
    \item \textbf{Single-Model Multi-Rollout Verification.} A single model performs multiple independent reasoning runs on the same question, each producing a candidate answer under all constraint conditions. A question passes only when all runs converge on the same entity that is consistent with the ground truth; otherwise, it is discarded.

    \item \textbf{Cross-Model Validation.} Multiple heterogeneous agentic models independently verify the same question in parallel, mitigating model-specific biases and knowledge blind spots. A question enters the final dataset only when all models reach consistent and correct conclusions.

    \item \textbf{Data Recovery.} Search API failures or parsing errors may cause valid samples to fail unexpectedly. Such samples are re-verified separately, and those that pass are retained in the final dataset.
\end{itemize}

\subsection{Difficulty Analysis}
\label{sec:difficulty}

Through the above pipeline, we synthesize tens of thousands of medical long-horizon deep search tasks. Subjective assessment based solely on surface question complexity fails to objectively reflect actual task difficulty. To this end, we quantitatively validate the difficulty of the synthesized data using the distribution of tool-call rounds as a proxy indicator, where more rounds indicate higher demands on multi-step planning and iterative retrieval.

\begin{figure}[h]
    \centering
    \includegraphics[width=\linewidth]{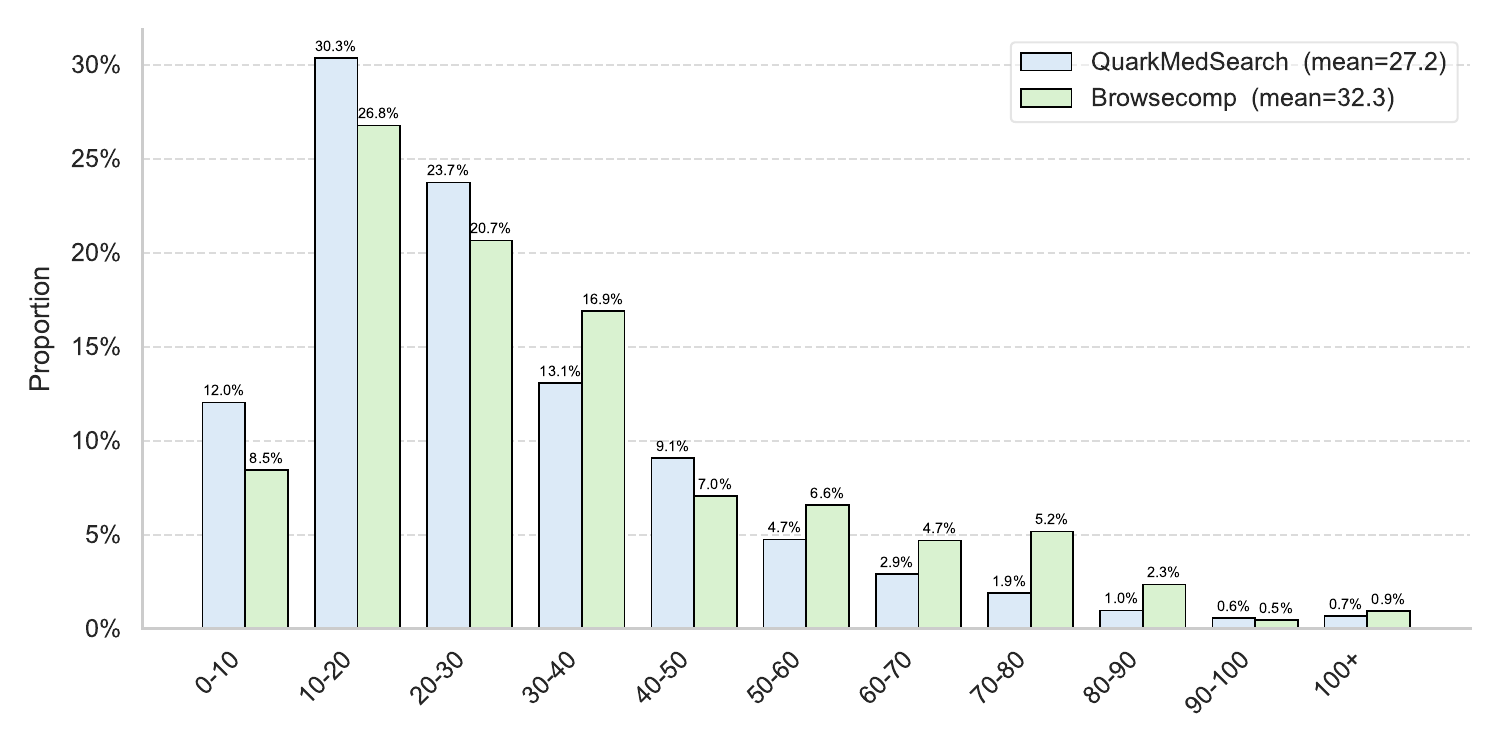}
    \caption{Comparison of the distribution of tool-call counts between QuarkMedSearch and BrowseComp.}
    \label{fig:difficulty}
\end{figure}

We compare a sampled subset of 500 questions from our synthesized dataset against BrowseComp, using Tongyi DeepResearch as the evaluation agent because of its high normal termination rate within a 128K context window. To ensure a fair comparison, we retain only samples for which Tongyi DeepResearch answers correctly at least once across multiple rollouts. As shown in Figure~\ref{fig:difficulty}, the average number of tool calls for our dataset and BrowseComp is similar (27.2 vs. 32.3), and the two datasets exhibit similar distribution shapes, indicating that the synthesized dataset achieves a comparable level of task difficulty. These results suggest that our data captures the core challenges of deep search scenarios, including long-horizon planning, multi-source evidence integration, and iterative retrieval, making it directly applicable to SFT trajectory synthesis or RL training.

\section{Training Recipe}

\begin{figure}[h]
    \centering
    \includegraphics[width=\linewidth]{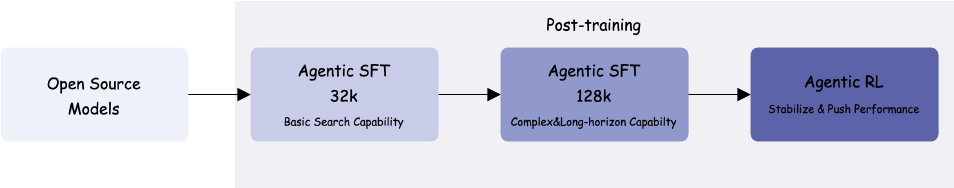}
    \caption{Post-training stages for QuarkMedSearch.}
    \label{fig:train_recipe}
\end{figure}

QuarkMedSearch is built on Tongyi DeepResearch 30B-A3B. We choose this backbone for two reasons. First, it already provides strong general capabilities in complex reasoning and multi-turn tool use. Second, it reflects a broader trend in the field: high-performing agentic foundation models are increasingly used as backbones for domain-specific adaptation. The overall training pipeline consists of two stages, as illustrated in Figure~\ref{fig:train_recipe}.

The first stage consists of two-phase SFT training. Trajectories are obtained via rejection sampling, with the shortest correct trajectory selected for each task. Training progresses from 32K tokens, which establishes basic tool-call conventions and search-reasoning patterns, to 128K tokens, which teaches the model to leverage long-horizon historical information and adjust its strategy in a timely manner; 20\% of short trajectories are retained for stability. The second stage is RLVR, in which boundary samples of moderate difficulty are selected to further improve the model's deep search capabilities through continuous environment interaction under a strict reward mechanism.

\subsection{Trajectory Synthesis and Data Filtering}

\paragraph{Trajectory Synthesis.}
Based on the medical deep search long-horizon tasks synthesized in Section~\ref{sec:Medical Deep Search Task Synthesis}, we employ multiple frontier agentic models to perform rollouts on each query. This multi-model design serves two purposes: (1) different models exhibit distinct patterns in intent planning, tool invocation, and information integration, providing diverse strategy coverage as a behavioral foundation for subsequent RL training; (2) from all rollouts, we select the correct trajectory with the shortest action sequence, encouraging the model to complete retrieval reasoning concisely rather than relying on redundant search.

\paragraph{Trajectory Filtering.}
We filter trajectories from two perspectives. For rule-based filtering, we remove trajectories exhibiting erroneous actions or format anomalies in tool invocations, repeated or semantically similar search queries and brute-force search without substantive reasoning (detected via n-gram overlap), and circular reasoning or non-terminating iterations. For rubric-based filtering, we evaluate long-trajectory samples (32K–128K tokens) across four dimensions: intent planning, tool invocation, information integration, and factual rigor (Table~\ref{tab:rubrics}), where factual rigor is a hard constraint that assigns a score of 0 whenever any factual error is detected. For medical QA data, information integration is further assessed in terms of medical terminology standardization, source authority, and timeliness.

\begin{table}[h]
\centering
\caption{Multi-Dimensional Rubrics for Trajectory Evaluation}
\label{tab:rubrics}
\begin{tabular}{lp{12cm}}
\toprule
\textbf{Dimension} & \textbf{Scoring Criteria} \\
\midrule
Intent Planning & Accuracy of problem understanding and solution path planning; requires clear decomposition of complex problems. \\
\addlinespace
Tool Invocation & Parameter accuracy, query construction quality, and fault tolerance; requires precise and efficient invocation without redundancy. \\
\addlinespace
Information Integration & Extraction of key information, strategy adjustment upon failed retrieval, and (for medical QA) terminology standardization, source authority and timeliness. \\
\addlinespace
Factual Rigor & Hard constraint: any factual error results in a score of 0. \\
\bottomrule
\end{tabular}
\end{table}

\subsection{Two-Phase SFT Training Strategy}
QuarkMedSearch adopts a two-phase SFT strategy to progressively transition the model from basic search behaviors to complex long-horizon reasoning, with observation portions masked from gradient updates in both phases. The first phase trains on short trajectories ($\leq$32K tokens) to establish basic tool-call conventions and the foundational ``think--search--observe--summarize'' framework, prioritizing format compliance and behavioral stability. The second phase fine-tunes from the first-phase checkpoint using long trajectories (32K–128K tokens) filtered by rubrics, enabling multi-step planning and iterative reasoning essential for medical scenarios such as cross-literature evidence synthesis, differential diagnosis, and drug interaction evaluation. To prevent capability degradation on shorter tasks, approximately 20\% of short samples ($\leq$32K tokens) are retained in the second-phase data.

In both phases, open-source general-purpose deep search trajectories are incorporated to prevent degradation of general capabilities. Low-quality or insufficiently difficult samples are augmented via the pipeline in Section~\ref{sec:Medical Deep Search Task Synthesis}, including multi-hop fact chain construction and entity obfuscation, and subjected to the same filtering and rubrics-based selection to ensure consistency with medical domain data in trajectory quality and strategy diversity.

\subsection{RL Training Algorithm}
\label{sec:RL Training Algorithm}

Following SFT, QuarkMedSearch applies reinforcement learning with verifiable rewards (RLVR) to further optimize search strategies and reasoning decisions through environment interaction. We adopt GRPO~\citep{shao2024deepseekmath}, which samples a group of trajectories per question, computes rewards, and normalizes within the group to derive relative advantages for policy updates. Following DAPO~\citep{yu_dapo_2025}, we remove the KL divergence term and adopt a token-level policy loss:

\begin{equation}
\mathcal{J}(\theta) = -\frac{1}{\sum_{i=1}^{G}|o_i|}
\sum_{i=1}^{G}\sum_{t=1}^{|o_i|}
\min\left(
    r_{i,t}(\theta)\,\hat{A}_{i,t},\;
    \text{clip}\!\left(r_{i,t}(\theta),\, 1-\varepsilon_{\text{low}},\, 1+\varepsilon_{\text{high}}\right)\hat{A}_{i,t}
\right)
\end{equation}
where $r_{i,t}(\theta)$ is the importance sampling ratio. Since we adopt strict on-policy training, the importance sampling ratio is always 1.0:

\begin{equation}
r_{i,t}(\theta) = \frac{\pi_\theta(o_{i,t} \mid q,\, o_{i,<t})}{\pi_{\theta_{\text{old}}}(o_{i,t} \mid q,\, o_{i,<t})}
\end{equation}
The within-group normalized advantage estimate $\hat{A}_{i,t}$ is obtained by normalizing the final reward $R_i$ of each trajectory using the within-group mean and standard deviation:

\begin{equation}
\hat{A}_{i,t} = \frac{R_i - \mathrm{mean}\!\left(\{R_j\}_{j=1}^{G}\right)}{\mathrm{std}\!\left(\{R_j\}_{j=1}^{G}\right)}
\end{equation}
We adopt Truncated Importance Sampling (TIS)~\citep{yao2025rollouttraining} and Route Replay (R2)~\citep{zheng_stabilizing_2025} to mitigate the train-inference inconsistency problem.

\paragraph{Reward Design.}
Designing the reward function is non-trivial, particularly in balancing format rewards and correctness rewards. During RL training, models tend to find shortcuts. Rather than genuinely improving search strategies and exploration accuracy, they learn to forcibly produce an answer within the context window, even if it is incorrect, simply to obtain format rewards. This leads to reward hacking. To address this issue, QuarkMedSearch adopts a stricter reward mechanism, defined as follows:
\begin{equation}
R = \begin{cases}
1.0 & \text{if } \textit{format\_score} = 1 \wedge \textit{correct\_score} = 1 \\
0.8 & \text{if } \textit{format\_score} = 0 \wedge \textit{correct\_score} = 1 \\
0   & \text{otherwise}
\end{cases}
\end{equation}
Specifically, a reward is granted only when the answer is correct, with format compliance determining the reward magnitude. This design compels the model to improve its search and reasoning capabilities, while simultaneously encouraging adherence to output format, thereby eliminating the incentive for reward hacking.

\paragraph{RL Dataset Construction.}
A high-quality RL dataset is a prerequisite for ensuring the effectiveness of training signals. We apply difficulty filtering to the dataset, removing samples that are either too easy or too difficult, as both extremes lead to reward saturation or reward sparsity, resulting in vanishing gradient signals and preventing effective policy learning. A portion of general-domain data is also mixed into the RL dataset to prevent degradation of general capabilities.

\paragraph{RL Training Environment and Framework.}
For the training environment, we construct two complementary search environments. The first is an \textbf{offline environment}: a local vector index is built from the full 2024 Wikipedia dump and professional medical literature, with accompanying local retrieval and virtual webpage reading tools, used for early-stage algorithm verification and rapid iteration. The second is an \textbf{online environment}: connected to the Google Search API and an in-house medical vertical search engine, with an online reading engine for real-time webpage content fetching, serving as the primary training environment for core experiments. For the training framework, rollout sequences in deep search tasks tend to be long, and a small number of extremely long sequences can introduce significant training bubbles in synchronous training, severely blocking overall progress. To address this, we implement an asynchronous rollout and reward computation pipeline. Each worker independently executes rollouts and returns results asynchronously, reducing the need to wait for the slowest sample. This effectively minimizes training bubbles caused by long-tail sequences and significantly improves overall training throughput.

\section{QuarkMedSearch Benchmark Construction}
\label{sec:QuarkMedSearchBenchmarkConstruction}

\paragraph{Data Sources.}
The construction of the QuarkMedSearch Benchmark is guided by two core principles: data quality and source diversity. Data is collected from two independent channels. The first is \textbf{open-source benchmark filtering}: we systematically screen high-difficulty, medically relevant questions from academically recognized benchmarks such as BrowseComp and HLE and incorporate them into the candidate pool. The second is \textbf{pipeline-synthesized data}: we apply the QuarkMedSearch data synthesis pipeline described in Section~\ref{sec:Medical Deep Search Task Synthesis} to generate a set of high-difficulty medical multi-hop reasoning questions. To prevent benchmark contamination, all collected data is strictly isolated from the SFT and RL training sets at the entity level, ensuring that no key entity in the benchmark appears among the core entities in the SFT and RL training data.

\paragraph{Expert Human Verification.}
We invite medical experts to perform item-by-item manual review of all candidate samples across three dimensions. (1) \textbf{Linguistic clarity}: each question must be grammatically correct and semantically unambiguous, leaving no room for misinterpretation that could yield multiple valid ground truth answers. (2) \textbf{Question difficulty}: each question must satisfy a minimum reasoning depth threshold, such that it cannot be answered from parametric memory alone by either medical experts or frontier LLMs, ensuring the benchmark effectively evaluates the model's ability to integrate medical knowledge with retrieval-based reasoning. (3) \textbf{Ground truth correctness and uniqueness}: all key conditions and their corresponding entities required for reasoning are rigorously verified to ensure that the reasoning trajectory converges unambiguously to a single correct answer.
\begin{figure}[h]
    \centering
    \includegraphics[width=\linewidth]{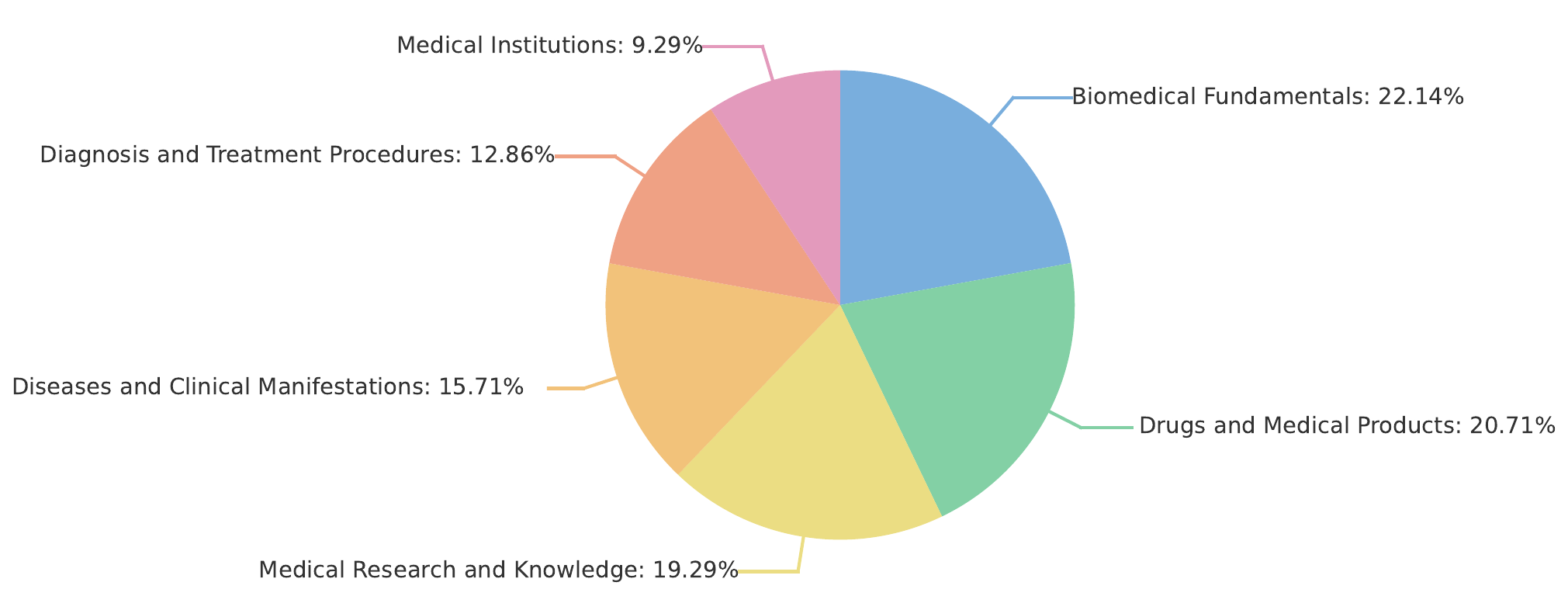}
    \caption{Distribution of Sample Categories in the QuarkMedSearch Benchmark.}
    \label{fig:category_dist}
\end{figure}

\paragraph{Data Filtering and Category Distribution.}
Each question is categorized according to the type of core entity involved in its reasoning process, resulting in six major categories (see Table~\ref{tab:categories}). To prevent systematic bias introduced by category imbalance, we apply stratified sampling based on these annotations: over-represented categories are pruned, while under-represented categories are supplemented with synthesized data. The final QuarkMedSearch Benchmark comprises high-quality medical long-horizon reasoning questions verified by medical experts, with the category distribution shown in Figure~\ref{fig:category_dist}.

\begin{table}[H]
\centering
\caption{Category Taxonomy of the QuarkMedSearch Benchmark}
\label{tab:categories}
\begin{tabular}{lp{9cm}}
\toprule
\textbf{Category} & \textbf{Key Entity Coverage} \\
\midrule
Biomedical Fundamentals & Anatomical structures, biochemical molecules, pathophysiological mechanisms, 
pathogenic microorganisms, etc. \\
\addlinespace
Drugs and Medical Products & Chemical drugs, active ingredients of herbal medicine, biologics, vaccines, 
medical devices, etc. \\
\addlinespace
Medical Research and Knowledge & Clinical trial findings, evidence-based medicine, medical statistics, 
diagnostic and treatment technical standards, etc. \\
\addlinespace
Diseases and Clinical Manifestations & Disease diagnosis and classification, clinical syndromes, symptoms and signs, 
key differential diagnosis points, etc. \\
\addlinespace
Clinical Procedures & Clinical laboratory tests, surgical procedures, treatment protocols, 
medication strategies and interventions, etc. \\
\addlinespace
Medical Institutions & Hospitals, pharmaceutical companies, medical device companies, 
medical academic associations, health regulatory agencies, etc. \\
\bottomrule
\end{tabular}
\end{table}

\section{Experiments}
\subsection{Experimental Setup}

\paragraph{Benchmarks.}
We evaluate QuarkMedSearch on several challenging benchmarks. For medical-domain evaluation, we use the QuarkMedSearch Benchmark introduced in Section~\ref{sec:QuarkMedSearchBenchmarkConstruction}, which is annotated and reviewed by domain experts and provides a fine-grained assessment of model performance in realistic medical scenarios. For general deep search, we additionally report results on BrowseComp-EN~\citep{wei_browsecomp_2025}, BrowseComp-ZH~\citep{zhou2025browsecompzh}, and Xbench DeepSearch~\citep{xbench}.

\paragraph{Baselines.}
We conduct systematic comparisons of QuarkMedSearch against mainstream search agent baselines. For general deep search capability evaluation, the baselines include: (1) state-of-the-art commercial closed-source models, such as Seed1.8~\citep{bytedance_seed_1_8_2025}, Gemini-3-Pro~\citep{gemini3pro2025}, Claude-4.5-opus~\citep{anthropic_claude_opus_4_5}, and GPT-5.2-Thinking-xhigh~\citep{singh2025openai}; (2) large-scale open-source models, including Kimi-K2.5~\citep{team_kimi_2026}, GLM-4.7~\citep{team_glm-45_2025}, GLM-5~\citep{team_glm-5_2026}, and DeepSeek-V3.2~\citep{deepseek-ai_deepseek-v32_2025}; and (3) open-source models of comparable parameter scale, including WebResearcher~\citep{qiao2025webresearcher}, WebSailorV2~\citep{li_websailor-v2_2025}, REDSearcher~\citep{chu_redsearcher_2026}, SMTL-100~\citep{chen_search_2026}, and Tongyi DeepResearch~\citep{team_tongyi_2025}. For medical domain capability evaluation, the comparison is further extended to large-scale general-purpose models such as Seed1.8~\citep{bytedance_seed_1_8_2025}, Kimi-K2.5~\citep{team_kimi_2026}, GLM-5~\citep{team_glm-5_2026}, DeepSeek-V3.2~\citep{deepseek-ai_deepseek-v32_2025}, and Qwen3.5-Plus~\citep{qwen3.5}, as well as open-source models of equivalent scale including Qwen3.5-35B-A3B~\citep{qwen3.5} and Qwen3.5-27B~\citep{qwen3.5}, so as to comprehensively assess the performance advantages of QuarkMedSearch over general-purpose models in medical vertical scenarios.

\paragraph{Evaluation Setup.}
To ensure fairness across all experiments, we adopt a unified configuration: the temperature is set to 0.6, the maximum context length is set to 128K tokens, and the maximum number of tool calls is set to 128, consistent with prior work. Unless otherwise specified, all benchmarks report Avg@3 as the default metric, where each question is sampled independently three times and the average score is used to smooth output randomness and obtain a more stable estimate of the model's true performance. It is worth noting that, in the evaluation on the QuarkMedSearch Benchmark, most existing open-source agentic base models do not publicly release their official evaluation scripts for search agents, resulting in considerable discrepancies across models in prompt design and evaluation environments. To address this, we apply a unified prompt template and evaluation environment consistent with that of QuarkMedSearch to all models involved in the comparison, ensuring the comparability of evaluation results and the reliability of the conclusions.

\paragraph{Training Details.}
Due to space constraints, detailed training configurations and hyperparameter settings are provided in Appendix A.

\subsection{Main Results}
\textbf{Medical Domain Capability Evaluation.} As shown in Table~\ref{tab:medical_results}, QuarkMedSearch demonstrates strong domain expertise on the QuarkMedSearch Benchmark, achieving state-of-the-art performance among all models of comparable parameter scale. Compared with Tongyi DeepResearch, the backbone of QuarkMedSearch, our model achieves substantial improvements, providing strong evidence for the effectiveness of our medical-oriented data construction and training strategy. More notably, QuarkMedSearch achieves performance on par with, or even better than, that of large-scale general-purpose models with significantly more parameters, while markedly outperforming open-source baselines of equivalent scale. These results demonstrate that, through targeted medical-domain data construction and training optimization, moderate-scale models can compete effectively with large-scale general-purpose models on specialized tasks.

\begin{table}[h]
\centering
\caption{Results on the QuarkMedSearch Benchmark.}
\label{tab:medical_results}
\begin{tabular}{llc}
\toprule
\textbf{Backbone} & \textbf{Size} & \textbf{QuarkMedSearch Benchmark} \\
\midrule
\rowcolor{gray!20}\multicolumn{3}{c}{\emph{\textbf{Large-scale Models}}} \\
Seed1.8 & - & 55.71 \\
Kimi-K2.5 & 1T-A32B & 56.74 \\
GLM-5 & 744B-A40B & 58.57 \\
DeepSeek-V3.2 & 671B-A37B & 57.14 \\
Qwen3.5-Plus & 397B-A17B & 51.42 \\
\rowcolor{gray!20}\multicolumn{3}{c}{\emph{\textbf{Medium-scale Models}}} \\
Qwen3.5-35B-A3B & 35B-A3B & 43.57 \\
Qwen3.5-27B & 27B & 44.28 \\
\rowcolor{gray!20}\multicolumn{3}{c}{\emph{\textbf{Ours}}} \\
Tongyi DeepResearch & 30B-A3B & 40.71 \\
\textbf{QuarkMedSearch} & \textbf{30B-A3B} & \textbf{55.71} \\
\bottomrule
\end{tabular}
\end{table}

\begin{table}[h]
\centering
\caption{Evaluation results on general deep search benchmarks. For commercial closed-source models, we directly adopt the best results from official reports. For open-source models, we report only the results without context management (w/o ctm) to ensure fair comparison, and include only those models that explicitly provide such results in the comparison among models of similar parameter scale. † denotes results obtained under our unified evaluation environment. Bold indicates the best result among models of similar parameter scale.}
\label{tab:general_results}
\begin{tabular}{llccc}
\toprule
\textbf{Backbone} & \textbf{Size} & \textbf{BrowseComp-EN} & \textbf{BrowseComp-ZH} & \textbf{Xbench DeepSearch} \\
\midrule
\rowcolor{gray!20}\multicolumn{5}{c}{\emph{\textbf{Commercial Closed-source Models}}} \\
Seed1.8 & - & \textbf{67.6} & \textbf{81.3} & - \\
Gemini-3-Pro & - & 59.2 & 66.8 & - \\
Claude Opus 4.5 & - & 57.8 & 62.3 & - \\
GPT-5.2 (xhigh) & - & 65.8 & 76.1 & - \\
\rowcolor{gray!20}\multicolumn{5}{c}{\emph{\textbf{Open-source Large-scale Models (w/o ctm)}}} \\
Kimi-K2.5 & 1T-A32B & 60.6 & - & - \\
GLM-4.7 & 355B-A32B & 52.0 & 67.5 & - \\
GLM-5 & 744B-A40B & \textbf{62.0} & \textbf{75.9} & - \\
DeepSeek-V3.2 & 671B-A37B & 51.4 & 65.0 & \textbf{71.0} \\
\rowcolor{gray!20}\multicolumn{5}{c}{\emph{\textbf{Open-source Medium-scale Models (w/o ctm)}}} \\
WebResearcher & 30B-A3B & 37.3 & 45.2 & 71.0 \\
WebSailorV2 & 30B-A3B & 35.3 & 44.1 & 73.7 \\
Tongyi DeepResearch & 30B-A3B & 43.4 & 46.7 & 75.0 \\
REDSearcher & 30B-A3B & 42.1 & 49.8 & - \\
SMTL-100 & 30B-A3B & \textbf{43.6} & - & \textbf{80.0} \\
\rowcolor{gray!20}\multicolumn{5}{c}{\emph{\textbf{Ours}}} \\
Base (Tongyi DeepResearch)$^\dagger$ & 30B-A3B & 42.67 & 45.40 & 74.0 \\
\textbf{QuarkMedSearch} & \textbf{30B-A3B} & \textbf{47.03} & \textbf{57.55} & \textbf{81.0} \\
\bottomrule
\end{tabular}
\end{table}
\textbf{General Deep Search Capability Evaluation.} As shown in Table~\ref{tab:general_results}, QuarkMedSearch remains competitive on general deep search benchmarks. Among open-source models of comparable scale, it outperforms same-scale baselines on BrowseComp-EN, BrowseComp-ZH, and Xbench DeepSearch. Relative to the base model Tongyi DeepResearch, gains are consistent across all three benchmarks and are especially notable on BrowseComp-ZH. These results suggest that our mixed training strategy strengthens medical-domain capability without materially sacrificing general search and reasoning ability.

\subsection{Analysis}

\subsubsection{Effectiveness of Two-Phase SFT}
To validate the necessity of the two-stage SFT strategy, we conduct an ablation study comparing the Stage-I model with its Stage-II counterpart further fine-tuned on long-context data. As shown in Table~\ref{tab:sft_ablation}, the Stage-I model already demonstrates basic deep search capabilities across all benchmarks. After second-stage fine-tuning, the model achieves gains of +5.96, +2.24, and +2.22 percentage points on BrowseComp-ZH, BrowseComp-EN, and the QuarkMedSearch Benchmark, respectively. These results confirm that long-context trajectory data is essential for adapting the model to complex multi-turn retrieval scenarios and improving reasoning coherence in ultra-long-sequence settings, and that short-context training alone is insufficient for the model to fully develop effective retrieval-planning capabilities.

\begin{table}[h]
\centering
\caption{Ablation study results of the two-stage SFT training strategy.}
\label{tab:sft_ablation}
\begin{tabular}{lccc}
\toprule
\textbf{Stage} & \textbf{QuarkMedSearch Benchmark} & \textbf{BrowseComp-EN} & \textbf{BrowseComp-ZH} \\
\midrule
Base(Tongyi DeepResearch) & 40.71 & 42.67 & 45.40 \\
+ 32K Stage I & 49.99 & 43.23 & 49.40 \\
+ 128K Stage II & 52.21 & 45.47 & 55.36 \\
\bottomrule
\end{tabular}
\end{table}

\subsubsection{RL Training Dynamics}

\begin{figure}[H]
    \centering
    \includegraphics[width=\linewidth]{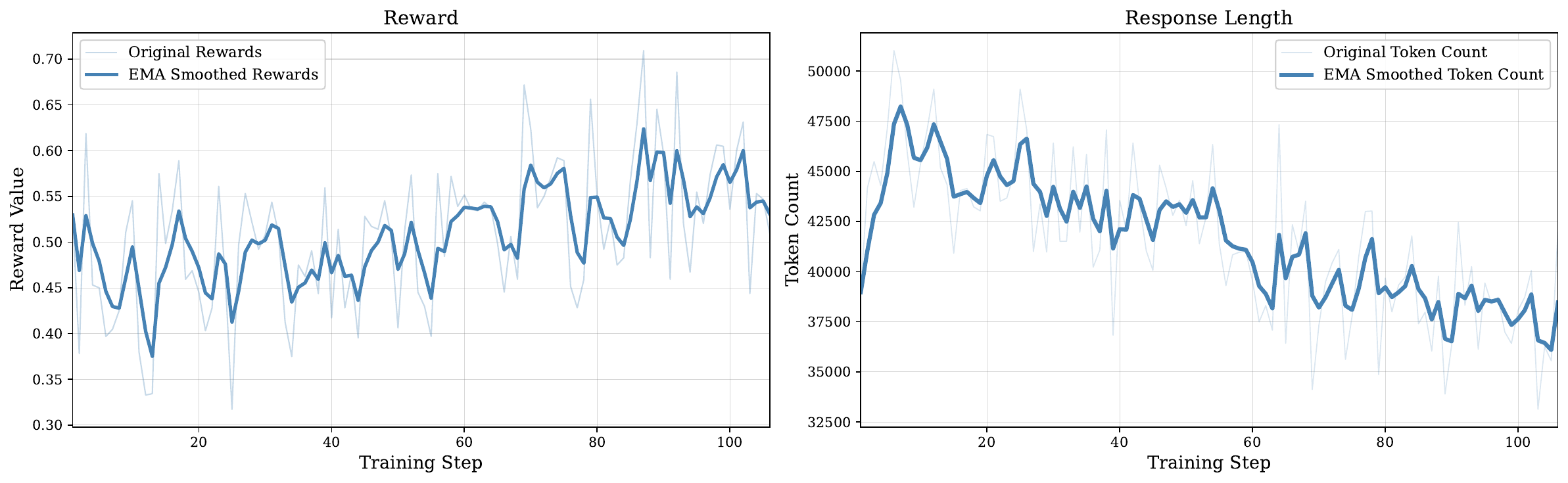}
    \caption{RL training dynamics: response length and reward over training steps.}
    \label{fig:rl_dynamics}
\end{figure}
As shown in Figure~\ref{fig:rl_dynamics}, response length decreases steadily over the course of RL training. Importantly, this compression does not come at the expense of correctness: the reward curve rises at the same time, indicating that the model becomes both more accurate and more concise. This pattern suggests that RL encourages more efficient search behavior by reducing redundant tool calls and favoring more targeted retrieval paths. To quantify this effect, we report termination rate and the average number of tool calls on BrowseComp-EN and BrowseComp-ZH, where termination rate is defined as the proportion of questions completed within the tool-call budget and context window. Table~\ref{tab:termination} shows that both metrics improve together during training. Relative to the two-stage SFT baseline, RL adds +2.19 and +1.56 percentage points on BrowseComp-ZH and BrowseComp-EN, respectively, confirming that RL elicits deeper search capability beyond what SFT alone provides.

\begin{table}[H]
\centering
\caption{Comparison of termination rate and average number of tool call rounds required to correctly answer questions before and after RL training. w/o RL and w/ RL correspond to the model states at the end of two-stage SFT and after RL convergence, respectively. ↑ means higher is better, ↓ means lower is better.}
\label{tab:termination}
\begin{tabular}{lcccc}
\toprule
\multirow{2}{*}{\textbf{Metric}} & \multicolumn{2}{c}{\textbf{BrowseComp-ZH}} & \multicolumn{2}{c}{\textbf{BrowseComp-EN}} \\
\cmidrule(lr){2-3}
\cmidrule(lr){4-5}
 & \textbf{w/o RL} & \textbf{w/ RL} & \textbf{w/o RL} & \textbf{w/ RL} \\
\midrule
Termination Rate $\uparrow$ & 0.676 & 0.725 & 0.580 & 0.635 \\
Avg. Tool Calls (Correct) $\downarrow$ & 28.09 & 24.01 & 40.03 & 37.12 \\
\bottomrule
\end{tabular}
\end{table}

\subsubsection{Impact of Reward Function Design}
\begin{figure}[h]
\centering
\includegraphics[width=\linewidth]{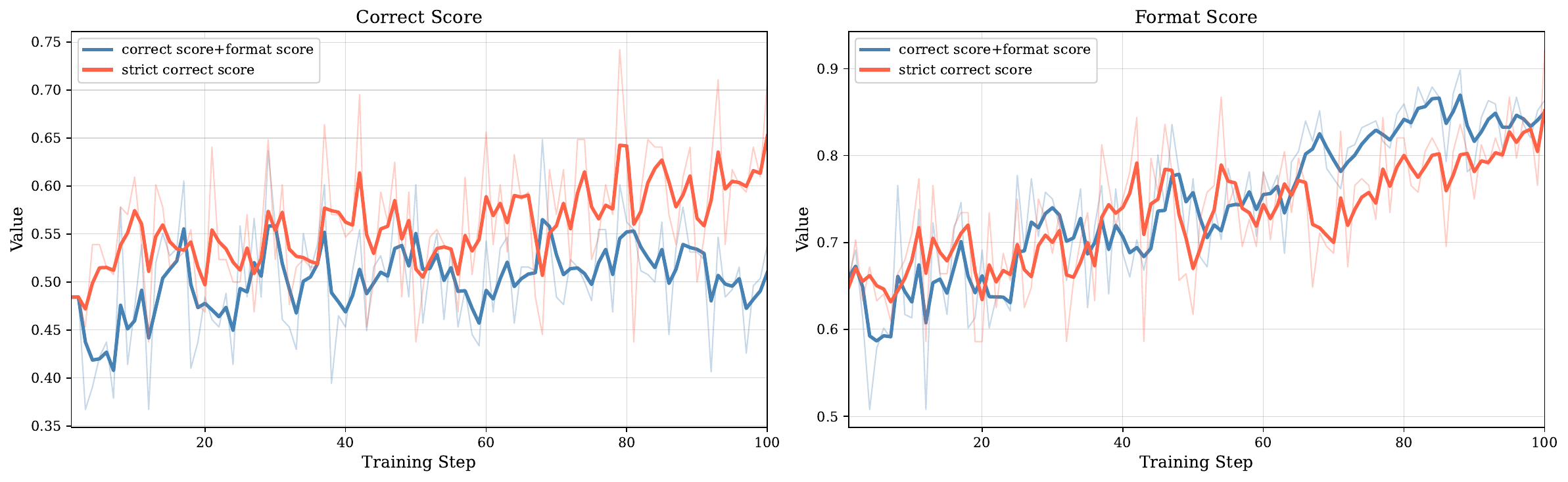}
\caption{Training dynamics of correct score and format score under two reward strategies: \textit{correct score + format score} (blue) and strict correct score (red).}
\label{fig:reward_design}
\end{figure}

Figure~\ref{fig:reward_design} compares the training dynamics under two reward design strategies. Under the \textit{correct score + format score} strategy, format score is incorporated into the total reward regardless of answer correctness. As observed, format score increases monotonically while correct score oscillates without a clear upward trend — a classic manifestation of reward hacking: even when the format reward constitutes only a small fraction of the total reward, the RL optimization process prioritizes this easier-to-exploit signal, suppressing the optimization of correctness.

To address this, QuarkMedSearch adopts the strict correct score strategy described in Section~\ref{sec:RL Training Algorithm}, where format score is incorporated only when correct score equals 1. As shown by the red curves, this strategy effectively eliminates reward hacking: correct score exhibits a stable upward trend with format score improving in tandem. These results highlight that RL optimization will always exploit the weakest signal in the reward function, and strictly gating auxiliary rewards is necessary to ensure stable and meaningful training.

\subsubsection{Effectiveness of Context Management}
\begin{table}[h]
\centering
\caption{Results with ctm on BrowseComp.}
\label{tab:context_management}
\begin{tabular}{lcc}
\toprule
\textbf{Method} & \textbf{BrowseComp-EN} & \textbf{BrowseComp-ZH} \\
\midrule
\rowcolor{gray!20}\multicolumn{3}{c}{\emph{\textbf{Large-scale Open-source Models (w/ ctm)}}} \\
Kimi-K2.5 & 74.9 & - \\
GLM-5 & 75.9 & 72.7 \\
Qwen3.5-Plus & 78.6 & 70.3 \\
DeepSeek-V3.2 & 67.6 & 65.0 \\
\rowcolor{gray!20}\multicolumn{3}{c}{\emph{\textbf{Open-source Models of Comparable Scale (w/ ctm)}}} \\
Qwen3.5-35B-A3B & 61.0 & 61.0 \\
Qwen3.5-27B & 62.1 & 69.5 \\
MiroThinker-V1.5 & 56.1 & 66.8 \\
REDSearcher & 57.4 & 58.2 \\
\rowcolor{gray!20}\multicolumn{3}{c}{\emph{\textbf{Ours}}} \\
QuarkMedSearch & 47.03 & 57.55 \\
QuarkMedSearch w/ Discard-all & 57.61 & 67.13 \\
\textbf{$\Delta$ Improvement} & \textbf{+10.58} & \textbf{+9.58} \\
\bottomrule
\end{tabular}
\end{table}

In deep search, repeated tool calls and accumulated Web content can cause the context to exceed the model's input window. As the context approaches this limit, long-horizon reasoning degrades sharply. In practice, context overflow often signals a deeper failure: the model has adopted a poor retrieval plan early on and continues to accumulate unproductive tool calls, a problem made worse by the variability of live Web content. Following DeepSeek-V3.2~\citep{deepseek-ai_deepseek-v32_2025} and Kimi-K2.5~\citep{team_kimi_2026}, we therefore adopt the \textbf{Discard-all} strategy. Once the context exceeds a preset threshold, all historical tool-call records are cleared and only the original question plus a minimal task description are retained, prompting the agent to restart reasoning within the remaining tool budget. This both creates an opportunity to correct a flawed retrieval plan and implicitly encourages more precise reasoning in fewer steps. As shown in Table~\ref{tab:context_management}, Discard-all improves BrowseComp-EN and BrowseComp-ZH by +10.58 and +9.58 points, respectively. Among open-source models of comparable scale, QuarkMedSearch w/ Discard-all matches or surpasses recently released baselines on both benchmarks while remaining competitive with much larger models.

\subsection{Application to Real-World Medical Deep Research Scenarios}

\begin{figure}[htbp]
    \centering
    \includegraphics[width=\linewidth]{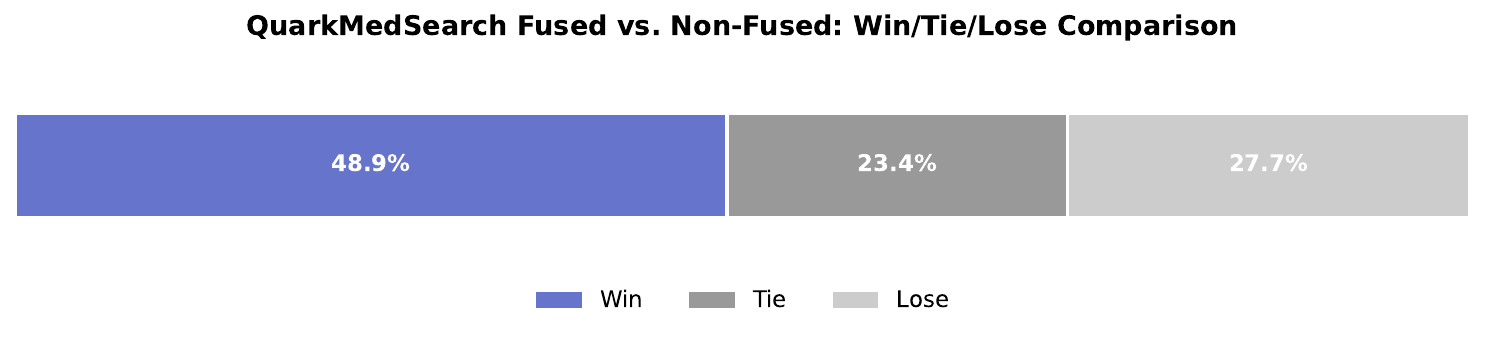}
    \caption{Win/Tie/Lose comparison between the fused model and the non-fused baseline, as evaluated by professional medical experts on medical Deep Research report generation.}
    \label{fig:gsb}
\end{figure}

In real-world medical Deep Research scenarios, user queries are predominantly open-ended and often lack definitive ground-truth answers, with users placing greater emphasis on the comprehensiveness and quality of the generated reports. This stands in sharp contrast to highly ambiguous puzzle-style tasks represented by benchmarks such as BrowseComp: the latter focuses on locating a single hidden answer, whereas the former emphasizes the systematic organization and in-depth interpretation of complex medical topics. To investigate whether the long-horizon deep search capability samples constructed by the QuarkMedSearch pipeline can generalize to real-world medical Deep Research scenarios, we explore a simple yet direct fusion strategy: mixing long-answer samples from real medical Deep Research scenarios with long-horizon deep search capability samples during training, without any additional task adaptation or structural modification. To evaluate this approach, we invited professional medical experts to conduct human evaluations of the model-generated deep research reports. Preliminary results indicate that the fused model achieves a G:S:B ratio of 48.9\%\,:\,23.4\%\,:\,27.7\% relative to the non-fused baseline, with the fused model preferred or tied in 72.3\% of cases (see Figure~\ref{fig:gsb}). These results suggest that, through simple sample fusion, the model can transfer, at least to some extent, the core capabilities acquired during long-horizon deep search training—including intent understanding, tool-call planning, and reflection-and-verification mechanisms—to real-world medical Deep Research report generation. The observed commonality between the two task types provides encouraging evidence and a useful reference point for future efforts to systematically apply deep search capabilities to real-world medical Deep Research scenarios.

\section{Related Works}
\paragraph{Agentic Deep Search Systems.}
Effectively integrating agentic intelligence into strong foundation-model capabilities has become a dominant research direction in the pursuit of artificial general intelligence (AGI). Early progress in this field was primarily driven by commercial systems centered on deep search and deep research. Mainstream commercial deep research systems such as OpenAI Deep Research~\citep{openai2025deepresearch}, Claude Research~\citep{anthropic2025multi_agent_research}, and Kimi-Researcher~\citep{moonshotai2025kimiresearcher} combine powerful proprietary foundation models with multi-step Web exploration, plan refinement, and long-context memory, continuously pushing the capability frontier of the field. As mainstream closed-source models such as Claude Opus 4.6~\citep{anthropic_claude_opus_4_6_2026}, Gemini 3.0~\citep{gemini3pro2025}, and GPT-5~\citep{singh2025openai} increasingly pursue general agentic intelligence, large-scale open-source models including Kimi-K2.5~\citep{team_kimi_2026}, DeepSeek-V3.2~\citep{deepseek-ai_deepseek-v32_2025}, the Qwen3.5 series~\citep{qwen3.5}, GLM-5~\citep{team_glm-5_2026}, and MiniMax-M2.5~\citep{minimax2026m25} have followed suit. Deep search, as a core component of agentic intelligence, has continued to make significant progress on benchmarks such as BrowseComp~\citep{wei_browsecomp_2025}, GAIA~\citep{mialon_gaia_2023}, and HLE~\citep{phan2025humanity}. Meanwhile, works such as MiroThinker~\citep{team_mirothinker_2025}, REDSearcher~\citep{chu_redsearcher_2026}, and Tongyi DeepResearch~\citep{team_tongyi_2025} show that, through targeted data construction and training optimization, models with moderate parameter counts can also achieve highly competitive results on the aforementioned benchmarks, further lowering the deployment barrier for high-performance deep search systems. However, these works are generally oriented toward general-purpose domains, and how to effectively transfer agentic deep-search capabilities to specialized scenarios remains an important direction that has yet to be fully addressed.

\paragraph{Deep Search Data Synthesis Pipelines.}
High-quality training data is the foundation of strong deep search agents, but collecting large-scale multi-step tool trajectories is prohibitively expensive. This has driven rapid progress in automated data-synthesis pipelines. Existing approaches can be divided into two broad categories. The first uses knowledge-graph-based generation, as in WebSailor~\citep{li_websailor_2025}, DeepDive~\citep{lu_deepdive_2025}, and InfoAgent~\citep{zhang_infoagent_2025}, which construct complex QA pairs by sampling paths over graphs. The second follows progressive difficulty scaling, as in WebExplorer~\citep{liu_webexplorer_2025}, which expands simple seed questions into long-horizon reasoning tasks with multi-step logical structure. TaskCraft~\citep{shi2025taskcraft} expands atomic tasks along both depth and breadth and introduces incremental verification to improve data quality and controllability. SMTL~\citep{chen_search_2026} instead extracts cross-domain URLs from real Web trajectories, parses them into raw corpora, converts the corpora into entity-relation graphs with LightRAG~\citep{guo2024lightrag}, and then synthesizes controllable multi-hop questions through random-walk subgraph extraction and iterative verification.
\paragraph{Deep Search in Medical Domain.}
Medical intelligence has long been one of the most important application directions in artificial intelligence, attracting sustained attention from both academia and industry~\citep{wu2026quarkmedbenchrealworldscenariodriven, xu2026quarkmedicalalignmentholistic}. Deep search in the medical domain is an important research direction that has emerged recently. Med-R³~\citep{lu_med-r3_2026} proposes a progressive reinforcement learning framework that decomposes the joint optimization of retrieval and reasoning into three stages and designs a multi-dimensional reward function encompassing entity-relation coverage and evidence-based medicine grading, validating the effectiveness of RL-driven retrieval-reasoning synergy in medical scenarios. MedResearcher-R1~\citep{yu_medresearcher-r1_2025} further incorporates multi-step retrieval and reasoning mechanisms, constructing training data through knowledge graph-based multi-hop path sampling, and achieves strong results on medical multi-step reasoning tasks. On the evaluation side, MedBrowseComp~\citep{chen_medbrowsecomp_2025} is the first benchmark dedicated to assessing agentic systems on multi-hop medical fact retrieval and synthesis from live, domain-specific knowledge bases, covering over 1,000 human-curated clinical questions and serving as an important testbed for evaluating the reliability of medical information seeking.

\section{Conclusion}

We present QuarkMedSearch, a framework for long-horizon deep search in the Chinese medical domain. To address the main weaknesses of prior work, we introduce a long-chain data-synthesis pipeline that integrates knowledge graphs with online search, combining long-tail entity sampling, multi-tool online exploration, entity obfuscation, and multi-model verification to produce challenging and reliable training data. We further adopt a post-training recipe that combines two-phase SFT with RLVR while mixing general-domain and medical data throughout training to strengthen medical expertise without eroding general capability. Experiments show that QuarkMedSearch achieves state-of-the-art performance on the QuarkMedSearch Benchmark while remaining competitive on general deep search benchmarks. We hope this work provides a reproducible reference for transferring agentic capability into specialized domains and helps advance Chinese medical AI toward stronger autonomous planning and deeper reasoning.

\clearpage

\bibliography{iclr2026_conference}
\bibliographystyle{iclr2026_conference}

\appendix
\clearpage

\section{Training Details}
QuarkMedSearch is built upon Tongyi DeepResearch 30B-A3B. For SFT, we use a batch size of 128 with a learning rate of 1e-6. For RL training, we employ GRPO with 16 questions per step and 8 rollout samples per question, yielding a minibatch size of 128, with the learning rate fixed at 1e-6. Following DAPO, we apply clip-higher while omitting entropy and KL losses. To address inconsistency issues, we incorporate truncated importance sampling (TIS) and route replay (R2). During inference, we set temperature to 0.85, top-$p$ to 0.95, and maximum context length to 128K tokens; if the context limit is exceeded, the model rolls back to the previous turn and is forced to generate an answer. Both the visit-tool summarizer and the LLM-as-a-judge evaluator are instantiated with Qwen3-Plus.

\section{Case Study}
We present a representative case from the QuarkMedSearch Benchmark, in which the model successfully completed a long-horizon medical deep-search task through 29 tool calls, spanning knowledge domains including phototherapy physics, DNA repair biology, hereditary hematology, clinical phenotyping, and real-time literature retrieval. Since most examples in the QuarkMedSearch Benchmark are in Chinese, we provide an English translation of this case for accessibility. Due to space constraints, tool responses are omitted.

\begin{enumerate}

\item \textbf{Task Decomposition.} Prior to any tool calls, the model decomposed the composite question into sequentially dependent sub-goals—from phototherapy wavelength-to-technology mapping, to endonuclease complementation group assignment, to corresponding-author affiliation retrieval—while maintaining a global problem structure to prevent local deadlocks from derailing overall reasoning.

\item \textbf{Hybrid Knowledge Utilization.} The model organically combined internal knowledge with external retrieval, directly applying well-established facts (e.g., ``pyrimidine dimers are repaired by NER,'' ``XPF is the 5$'$ endonuclease of NER'') while concentrating tool calls on genuinely uncertain nodes (e.g., the gene corresponding to FA complementation group Q; the institutional affiliation of the corresponding author of a 2025 DEK study), thereby improving overall retrieval efficiency.

\item \textbf{Adaptive Query Reformulation.} Upon five consecutive failed searches combining ``chromatin architecture protein + Fanconi anemia + 2025,'' the model iteratively adjusted its strategy—attempting specific protein names (CTCF, Cohesin) before retreating to the functional description ``chromatin + novel therapeutic strategies''—ultimately retrieving the target DEK inhibition study published in the \textit{Journal of Experimental Medicine} (2025).

\item \textbf{Error Detection and Recovery.} After incorrectly mapping FA complementation group Q to FAN1 and issuing several searches under this false assumption, the model detected the contradiction and re-entered via an alternative path—``phototherapy wavelength $\rightarrow$ NER pathway $\rightarrow$ XPF endonuclease''—ultimately confirming FANCQ $\equiv$ XPF/ERCC4. This ``deadlock $\rightarrow$ pivot $\rightarrow$ rejoin'' pattern reflects robust recovery from local reasoning impasses.

\item \textbf{Multi-Source Cross-Validation.} At critical nodes, the model proactively issued independent follow-up queries rather than accepting single-source conclusions. For instance, after obtaining ``FANCQ = XPF'' from a review, the model issued an explicit confirmatory search; the characterization of DEK as a ``chromatin architecture protein'' was similarly verified independently.

\item \textbf{Semantic Disambiguation.} In the final stage, the model retrieved two concurrent 2025 records for the Second Affiliated Hospital of Chongqing Medical University—selection into a flagship hospital construction program and approval of 51 NSFC grants. Rather than returning both indiscriminately, the model applied semantic understanding of ``honor program'' to distinguish a title-based designation from a funding-based award, precisely converging on the intended answer.

\end{enumerate}

\begin{tcolorbox}[breakable,title=Case Trajectory in the QuarkMedSearch Benchmark]
\textcolor{black}{\textbf{Question:} A phototherapy technique with a wavelength of approximately 300-plus nanometers is primarily used to treat cutaneous lymphoma. Its DNA damage mechanism involves the formation of covalent ring structures between adjacent pyrimidine bases. Repairing this damage requires a multi-protein excision repair system, one of whose endonuclease components simultaneously belongs to complementation group Q of a hereditary blood disorder. The core substrate protein monoubitiquitinated by the E3 ligase complex in which this endonuclease participates has a gene mutation that defines a subtype of blood disease. This subtype commonly presents with upper limb developmental abnormalities in the form of radial skeletal defects. The latest research (2025) found that an inhibitory strategy targeting a certain chromatin architectural protein can ameliorate this disease. What is the name of the honorary project obtained in 2025 by the main affiliated hospital of the medical school in Southwest China where the corresponding author of this study is located?}\\

\colorbox{gray!70}{%
    \parbox{\dimexpr\textwidth-2\fboxsep}{%
    \centering
        \textbf{Step 1}%
    }%
}
\textcolor{purple1}{\textbf{<think>} \\
I need to solve this problem step by step. First, let me break down the question: it concerns phototherapy for cutaneous lymphoma, where the DNA damage involves covalent cyclic structures between adjacent pyrimidine bases (likely thymine dimers), repaired by a multi-protein excision repair system. One endonuclease component belongs to the Q-type complementation group of a hereditary blood disorder. The core substrate protein ubiquitinated by the E3 ligase complex that this endonuclease participates in has a gene mutation that defines a subtype of blood disease characterized by upper limb developmental abnormalities and radial skeletal defects. The latest 2025 research found that an inhibitory strategy targeting a chromatin architectural protein can ameliorate this disease. The question asks what honor the primary affiliated hospital of the corresponding author's southwestern medical school received in 2025.\\
\\
First, I need to identify each molecular component, disease, author, and hospital. The DNA damage involves covalent cyclic structures between adjacent pyrimidine bases---i.e., UV-induced pyrimidine dimers---repaired by the nucleotide excision repair (NER) system. NER endonuclease components include XPG, XPF-ERCC1, etc. Then, which endonuclease belongs to the Q-type complementation group of a hereditary blood disorder? This likely refers to Fanconi anemia (FA), which has multiple complementation groups (FA-A through FA-Q and beyond). FA-Q corresponds to FAN1, which encodes a structure-specific endonuclease involved in DNA interstrand crosslink repair. \\
\\
Next, the E3 ligase complex in the Fanconi anemia pathway that performs monoubiquitination: the key E3 ligase is FANCL, which monoubiquitinates the core substrates FANCD2 and FANCI. A mutation in the gene encoding one of these core substrate proteins defines a subtype of blood disease with upper limb developmental abnormalities (radial skeletal defects), consistent with Fanconi anemia subtypes (e.g., FA-D2, caused by FANCD2 mutation), whose hallmarks include congenital bone marrow failure and radial/thumb anomalies. \\
\\
The 2025 study targets a chromatin architectural protein whose inhibition ameliorates this disease; the corresponding author is affiliated with a southwestern Chinese medical institution. Key southwestern medical centers include West China Hospital (Sichuan University), Chongqing Medical University affiliated hospitals, and Kunming Medical University. West China Hospital is the most prominent southwestern medical school affiliate.\\
\\
I will first confirm whether FAN1 is an NER endonuclease belonging to FA complementation group Q, and then identify the relevant disease subtype and the 2025 study.\\
\textbf{</think>}}\\
\textcolor{black}{\textbf{<tool\_call>} \\
\{"name": "search", "arguments": \{"query": ["FAN1 endonuclease Fanconi anemia complementation group Q", "Fanconi anemia upper limb abnormalities radial skeletal defects subtype", "2025 chromatin architectural protein inhibition Fanconi anemia corresponding author medical school southwestern"]\}\}
\\
\textbf{</tool\_call>}}\\

\colorbox{gray!70}{%
    \parbox{\dimexpr\textwidth-2\fboxsep}{%
    \centering
        \textbf{Step 2}%
    }%
}
\textcolor{purple1}{\textbf{<think>} \\
From the search results, upper limb skeletal defects in Fanconi Anemia (FA) are characterized by absence or hypoplasia on the radial side (e.g., radial ray defects). Regarding the FA-Q complementation group, I need to confirm if the corresponding gene is \textit{FAN1}. Search result 4 mentions "Fanconi anemia complementation group Q" but does not specify the gene. I will search to verify if \textit{FAN1} corresponds to group Q.\\
\textbf{</think>}}\\
\textcolor{black}{\textbf{<tool\_call>} \\
\{"name": "search", "arguments": \{"query": ["Fanconi anemia complementation group Q gene", "FAN1 gene Fanconi anemia type Q"]\}\} \\
\textbf{</tool\_call>}}\\

\colorbox{gray!70}{%
    \parbox{\dimexpr\textwidth-2\fboxsep}{%
    \centering
        \textbf{Step 3}%
    }%
}
\textcolor{purple1}{\textbf{<think>} \\
I need to definitively confirm the gene associated with complementation group Q. Previous search result 3 indicated that group Q is caused by a gene mutation but did not name the gene. I will perform a targeted search combining "FA-Q", "Fanconi anemia type Q", and "FAN1" to establish the link.\\
\textbf{</think>}}\\
\textcolor{black}{\textbf{<tool\_call>} \\
\{"name": "search", "arguments": \{"query": ["FA-Q Fanconi anemia gene FAN1", "FAN1 mutation Fanconi anemia complementation group Q"]\}\} \\
\textbf{</tool\_call>}}\\

\colorbox{gray!70}{%
    \parbox{\dimexpr\textwidth-2\fboxsep}{%
    \centering
        \textbf{Step 4}%
    }%
}
\textcolor{purple1}{\textbf{<think>} \\
I need to finalize the mapping between FA complementation groups and their genes. While groups A through W exist (totaling roughly 23 subtypes), the specific designation for group Q needs verification against standard nomenclature. The standard mapping is: FA-A (\textit{FANCA}), FA-B (\textit{FANCB}), ..., FA-D1 (\textit{BRCA2}/\textit{FANCD1}), ..., FA-J (\textit{BRIP1}/\textit{FANCJ}), etc. My hypothesis is that FA-Q corresponds to \textit{FAN1}, distinct from other groups like FA-D1 or FA-N. I will search specifically for "FAN1 Fanconi anemia type Q" to confirm that \textit{FAN1} is indeed the gene for complementation group Q.\\
\textbf{</think>}}\\
\textcolor{black}{\textbf{<tool\_call>} \\
\{"name": "search", "arguments": \{"query": ["FAN1 Fanconi anemia type Q confirmation", "FAN1 gene FA-Q complementation group"]\}\} \\
\textbf{</tool\_call>}}\\

\colorbox{gray!70}{%
    \parbox{\dimexpr\textwidth-2\fboxsep}{%
    \centering
        \textbf{Step 5}%
    }%
}
\textcolor{purple1}{\textbf{<think>} \\
Returning to the problem statement: "DNA damage mechanism involves covalent cyclic structures formed between adjacent pyrimidine bases." This describes pyrimidine dimers (e.g., thymine dimers), which are repaired by the Nucleotide Excision Repair (NER) system. Key endonucleases in NER include XPG (ERCC5) and XPF (ERCC4). I need to determine if any of these correspond to the "Group Q" of a hereditary blood disease. Previously, I hypothesized Fanconi Anemia (FA) Group Q might be \textit{FAN1}, but I may have conflated repair pathways. FA typically repairs interstrand crosslinks, while pyrimidine dimers are NER substrates. However, there is overlap. Let's re-examine the phototherapy clue: "wavelength around 300+ nm" used for cutaneous lymphoma. This likely refers to Narrowband UVB (NB-UVB, ~311 nm). I will search to confirm the specific phototherapy and its associated DNA damage type.\\
\textbf{</think>}}\\
\textcolor{black}{\textbf{<tool\_call>} \\
\{"name": "search", "arguments": \{"query": ["wavelength 300 nm phototherapy cutaneous lymphoma", "narrowband UVB 311 nm skin lymphoma treatment"]\}\} \\
\textbf{</tool\_call>}}\\

\colorbox{gray!70}{%
    \parbox{\dimexpr\textwidth-2\fboxsep}{%
    \centering
        \textbf{Step 6}%
    }%
}
\textcolor{purple1}{\textbf{<think>} \\
Search results confirm Narrowband UVB (~311 nm) is used for cutaneous T-cell lymphoma, causing pyrimidine dimers repaired by NER. The problem states an endonuclease in this repair system belongs to "Group Q" of a hereditary \textit{blood} disease. Xeroderma Pigmentosum (XP) has groups A-G and involves NER, but it is a skin disease, not primarily a blood disease. Fanconi Anemia (FA) is a hereditary blood disease with complementation groups up to W. Is there an FA group Q linked to an NER endonuclease? My previous assumption that FA-Q is \textit{FAN1} might be incorrect if \textit{FAN1} isn't an NER endonuclease. I need to explicitly search for the gene name associated with Fanconi Anemia Complementation Group Q.\\
\textbf{</think>}}\\
\textcolor{black}{\textbf{<tool\_call>} \\
\{"name": "search", "arguments": \{"query": ["Fanconi anemia complementation group Q gene name", "FANCQ gene mutation"]\}\} \\
\textbf{</tool\_call>}}\\

\colorbox{gray!70}{%
    \parbox{\dimexpr\textwidth-2\fboxsep}{%
    \centering
        \textbf{Step 7}%
    }%
}
\textcolor{purple1}{\textbf{<think>} \\
Search results indicate that Fanconi Anemia Group Q is caused by mutations in the \textit{FANCQ} gene. Now I need to identify the protein product of \textit{FANCQ} and its function. Specifically, is it an endonuclease involved in NER? If \textit{FANCQ} encodes an NER endonuclease like XPF or XPG, the puzzle pieces fit together. I will search for the relationship between \textit{FANCQ} and known NER endonucleases.\\
\textbf{</think>}}\\
\textcolor{black}{\textbf{<tool\_call>} \\
\{"name": "search", "arguments": \{"query": ["FANCQ protein function endonuclease", "FANCQ NER pathway"]\}\} \\
\textbf{</tool\_call>}}\\

\colorbox{gray!70}{%
    \parbox{\dimexpr\textwidth-2\fboxsep}{%
    \centering
        \textbf{Step 8}%
    }%
}
\textcolor{purple1}{\textbf{<think>} \\
Cross-validation step: Search results suggest \textit{FANCQ} is identical to \textit{XPF} (also known as \textit{ERCC4}), which is a structure-specific endonuclease essential for the 5' incision in NER. This confirms that the "Group Q" of the hereditary blood disease (Fanconi Anemia) corresponds to the \textit{XPF} gene. Now I must verify the next part of the query regarding the E3 ligase complex. The FA pathway involves an E3 ubiquitin ligase complex (containing FANCL) that monoubiquitinates core substrate proteins (FANCD2/FANCI). Mutations in these substrates define other FA subtypes (e.g., FA-D1/D2). The problem mentions a subtype characterized by radial ray defects (upper limb abnormalities). I need to confirm which FA subtype is most strongly associated with radial skeletal defects.\\
\textbf{</think>}}\\
\textcolor{black}{\textbf{<tool\_call>} \\
\{"name": "search", "arguments": \{"query": ["FANCQ XPF same gene", "FANCL E3 ligase FANCD2 monoubiquitination"]\}\} \\
\textbf{</tool\_call>}}\\

\colorbox{gray!70}{%
    \parbox{\dimexpr\textwidth-2\fboxsep}{%
    \centering
        \textbf{Step 9}%
    }%
}
\textcolor{purple1}{\textbf{<think>} \\
Final synthesis: The phototherapy is NB-UVB (~311 nm), causing pyrimidine dimers. Repair is via NER. The endonuclease XPF (encoded by \textit{ERCC4}) is also known as \textit{FANCQ}, linking it to Fanconi Anemia Group Q (a hereditary blood disease). The FA pathway uses the FANCL E3 ligase complex to monoubiquitinate FANCD2/FANCI. Mutations in genes encoding these core substrates or related proteins cause FA subtypes often presenting with radial ray defects (absence/hypoplasia of the radius and thumb). I will perform a final search to explicitly link Fanconi Anemia subtypes to radial skeletal defects to ensure the specific subtype mentioned in the problem (likely FA-D2 or general FA phenotype) aligns with the description.\\
\textbf{</think>}}\\
\textcolor{black}{\textbf{<tool\_call>} \\
\{"name": "search", "arguments": \{"query": ["Fanconi anemia radial ray defects upper limb abnormalities", "FANCD2 mutation clinical features radial defects"]\}\} \\
\textbf{</tool\_call>}}\\

\colorbox{gray!70}{%
    \parbox{\dimexpr\textwidth-2\fboxsep}{%
    \centering
        \textbf{Step 10}%
    }%
}
\textcolor{purple1}{\textbf{<think>} \\
Search results confirm that radial skeletal defects (absence or hypoplasia of the radius on the thumb side) are a hallmark of Fanconi Anemia (FA). The problem states that mutations in the gene encoding the "core substrate protein" monoubiquitinated by the E3 ligase complex define a specific FA subtype. In the FA pathway, the E3 ligase (FANCL complex) monoubiquitinates the FANCD2-FANCI heterodimer. Therefore, the core substrate is likely \textit{FANCD2} (defining FA-D2) or \textit{FANCI}. The next clue involves a 2025 study where inhibiting a "chromatin architecture protein" improves the disease. The corresponding author is from a medical school in Southwest China (e.g., Sichuan University/West China Hospital, Chongqing Medical University). I need to search for this specific 2025 research linking chromatin architecture proteins to Fanconi Anemia treatment.\\
\textbf{</think>}}\\
\textcolor{black}{\textbf{<tool\_call>} \\
\{"name": "search", "arguments": \{"query": ["2025 Fanconi anemia chromatin architecture protein inhibition", "2025 Fanconi anemia CTCF cohesin inhibitor study"]\}\} \\
\textbf{</tool\_call>}}\\

\colorbox{gray!70}{%
    \parbox{\dimexpr\textwidth-2\fboxsep}{%
    \centering
        \textbf{Step 11}%
    }%
}
\textcolor{purple1}{\textbf{<think>} \\
Previous searches did not yield the specific 2025 study from Southwest China. One result mentioned Wuhan University, which is not in the Southwest. I need to refine the search to target institutions in Southwest China, such as West China Hospital (Sichuan University) or Chongqing Medical University, combined with keywords like "Fanconi Anemia," "chromatin," and "2025."\\
\textbf{</think>}}\\
\textcolor{black}{\textbf{<tool\_call>} \\
\{"name": "search", "arguments": \{"query": ["2025 Fanconi anemia chromatin architecture West China Hospital", "2025 Fanconi anemia chromatin inhibitor Chongqing Medical University"]\}\} \\
\textbf{</tool\_call>}}\\

\colorbox{gray!70}{%
    \parbox{\dimexpr\textwidth-2\fboxsep}{%
    \centering
        \textbf{Step 12}%
    }%
}
\textcolor{purple1}{\textbf{<think>} \\
Still no direct hit. The term "chromatin architecture protein" might refer to specific complexes like Cohesin, Condensin, or proteins like CTCF. Since FA involves DNA repair and replication stress, Cohesin is a strong candidate. I will try searching specifically for "Cohesin" and "Fanconi Anemia" with a 2025 date limit and Southwest China authors. Alternatively, the study might be about "transcriptional stress" or "R-loops" mediated by architecture proteins.\\
\textbf{</think>}}\\
\textcolor{black}{\textbf{<tool\_call>} \\
\{"name": "search", "arguments": \{"query": ["2025 Fanconi anemia Cohesin inhibition study", "2025 Fanconi anemia chromatin structure protein Southwest China"]\}\} \\
\textbf{</tool\_call>}}\\

\colorbox{gray!70}{%
    \parbox{\dimexpr\textwidth-2\fboxsep}{%
    \centering
        \textbf{Step 13}%
    }%
}
\textcolor{purple1}{\textbf{<think>} \\
Let's broaden the search slightly to find recent papers (late 2024-2025) on Fanconi Anemia and chromatin regulators from Chinese institutions, then filter for Southwest locations. The key might be a specific finding about suppressing a protein to rescue FA phenotypes. I will search for general terms like "Fanconi anemia 2025 new therapy chromatin" and look for author affiliations.\\
\textbf{</think>}}\\
\textcolor{black}{\textbf{<tool\_call>} \\
\{"name": "search", "arguments": \{"query": ["Fanconi anemia 2025 new therapy chromatin regulator China", "Fanconi anemia 2025 paper corresponding author Sichuan Chongqing"]\}\} \\
\textbf{</tool\_call>}}\\

\colorbox{gray!70}{%
    \parbox{\dimexpr\textwidth-2\fboxsep}{%
    \centering
        \textbf{Step 14}%
    }%
}
\textcolor{purple1}{\textbf{<think>} \\
Hypothesis refinement: The "chromatin architecture protein" could be related to the resolution of R-loops or transcription-replication conflicts, where Cohesin or Topoisomerases play a role. If the specific 2025 paper is elusive via generic keywords, I will focus on identifying major hematology departments in Southwest China (West China Hospital, Children's Hospital of Chongqing Medical University) and their recent high-profile publications regarding FA. Once the specific paper and hospital are identified, I will search for the "honorary project" or award this hospital received in 2025.\\
\textbf{</think>}}\\
\textcolor{black}{\textbf{<tool\_call>} \\
\{"name": "search", "arguments": \{"query": ["West China Hospital Hematology 2025 Fanconi anemia research", "Chongqing Medical University Attached Hospital 2025 honor project hematology"]\}\} \\
\textbf{</tool\_call>}}\\

\colorbox{gray!70}{%
    \parbox{\dimexpr\textwidth-2\fboxsep}{%
    \centering
        \textbf{Step 15}%
    }%
}
\textcolor{purple1}{\textbf{<think>} \\
Previous searches for "chromatin architecture protein" and "Fanconi Anemia" in Southwest China were inconclusive. I need to refine the keywords. Perhaps the specific protein name or a different phrasing like "new strategy 2025 Fanconi Anemia chromatin" will yield results. If the disease identification (FA) or the Q-group (XPF/FANCQ) was incorrect, I might need to re-evaluate, but the clinical symptoms (radial defects) strongly point to FA. Let's try a broader search on recent therapeutic strategies involving chromatin in FA.\\
\textbf{</think>}}\\
\textcolor{black}{\textbf{<tool\_call>} \\
\{"name": "search", "arguments": \{"query": ["2025 Fanconi Anemia chromatin new strategy", "2025 Fanconi Anemia chromatin architecture protein inhibition China"]\}\} \\
\textbf{</tool\_call>}}\\

\colorbox{gray!70}{%
    \parbox{\dimexpr\textwidth-2\fboxsep}{%
    \centering
        \textbf{Step 16}%
    }%
}
\textcolor{purple1}{\textbf{<think>} \\
Still no direct hits. Let's double-check the phototherapy wavelength and the disease classification. The problem mentions "wavelength around 300+ nm" for cutaneous lymphoma. While NB-UVB (311 nm) is common, could it be 308 nm excimer laser? Also, is there any other hereditary blood disease with a "Group Q"? Fanconi Anemia remains the strongest candidate due to the radial ray defects. Let's re-verify the phototherapy specifics to ensure the DNA damage type (pyrimidine dimers) and repair pathway (NER) are correct, as this links to XPF/FANCQ.\\
\textbf{</think>}}\\
\textcolor{black}{\textbf{<tool\_call>} \\
\{"name": "search", "arguments": \{"query": ["cutaneous lymphoma phototherapy wavelength 300 nm", "308 nm excimer laser cutaneous T-cell lymphoma"]\}\} \\
\textbf{</tool\_call>}}\\

\colorbox{gray!70}{%
    \parbox{\dimexpr\textwidth-2\fboxsep}{%
    \centering
        \textbf{Step 17}%
    }%
}
\textcolor{purple1}{\textbf{<think>} \\
Re-focusing on the 2025 study clue: "Inhibition of a chromatin architecture protein improves the disease." Could the protein be CTCF? Or perhaps a less common one? Let's try searching specifically for "CTCF" and "Fanconi Anemia" with a 2025 date filter. If the previous assumption about the protein name was too generic, specifying potential candidates might help locate the paper from Southwest China.\\
\textbf{</think>}}\\
\textcolor{black}{\textbf{<tool\_call>} \\
\{"name": "search", "arguments": \{"query": ["2025 CTCF Fanconi Anemia study", "2025 chromatin architecture protein inhibitor Fanconi Anemia"]\}\} \\
\textbf{</tool\_call>}}\\

\colorbox{gray!70}{%
    \parbox{\dimexpr\textwidth-2\fboxsep}{%
    \centering
        \textbf{Step 18}%
    }%
}
\textcolor{purple1}{\textbf{<think>} \\
Breakthrough idea: Search results hint at a specific protein "DEK". A paper titled "Inhibition of DEK restores hematopoietic stem cell function in Fanconi anemia" published in \textit{J Exp Med} (March 2025) fits the description perfectly. DEK is known as a chromatin architecture protein. Now I need to confirm the corresponding author's affiliation. Is it a medical school in Southwest China (e.g., Chongqing Medical University)? I will search for the authors of this specific paper.\\
\textbf{</think>}}\\
\textcolor{black}{\textbf{<tool\_call>} \\
\{"name": "search", "arguments": \{"query": ["Inhibition of DEK restores hematopoietic stem cell function in Fanconi anemia authors", "DEK Fanconi Anemia 2025 corresponding author Chongqing"]\}\} \\
\textbf{</tool\_call>}}\\

\colorbox{gray!70}{%
    \parbox{\dimexpr\textwidth-2\fboxsep}{%
    \centering
        \textbf{Step 19}%
    }%
}
\textcolor{purple1}{\textbf{<think>} \\
Confirmation step: Search results indicate the corresponding author is Prof. Deng Jianchuan from the Second Affiliated Hospital of Chongqing Medical University (Southwest China). The study confirms DEK is a chromatin architecture protein and its inhibition helps FA. Now, the final task is to find the "honorary project" or award this hospital received in 2025. I will search for news regarding awards or honors for the Second Affiliated Hospital of Chongqing Medical University in 2025.\\
\textbf{</think>}}\\
\textcolor{black}{\textbf{<tool\_call>} \\
\{"name": "search", "arguments": \{"query": ["Second Affiliated Hospital of Chongqing Medical University 2025 honor project", "Chongqing Medical University Attached Hospital 2025 award news"]\}\} \\
\textbf{</tool\_call>}}\\

\colorbox{gray!70}{%
    \parbox{\dimexpr\textwidth-2\fboxsep}{%
    \centering
        \textbf{Step 20}%
    }%
}
\textcolor{purple1}{\textbf{<think>} \\
Now looking at search result 1: "The Second Affiliated Hospital of Chongqing Medical University Approved for Chongqing Integrated Traditional Chinese and Western Medicine 'Flagship' Hospital Construction Project" (December 4, 2025), and search result 2: "51 Items! The Second Affiliated Hospital of Chongqing Medical University Achieves a Record High in National Natural Science Foundation Grants in 2025" (August 27, 2025). I need to confirm which one qualifies as an "honorary project"? The question asks: "What honorary project was obtained in 2025?".
Let's return to the news link of that paper: Search "Inhibition of DEK Fanconi Anemia Chongqing Medical University corresponding author" to confirm whether the corresponding author's hospital is indeed the Second Affiliated Hospital of Chongqing Medical University.\\
\textbf{</think>}}\\
\textcolor{black}{\textbf{<tool\_call>} \\
\{"name": "search", "arguments": \{"query": ["Inhibition of DEK Fanconi Anemia Chongqing Medical University corresponding author"]\}\} \\
\textbf{</tool\_call>}}\\

\colorbox{gray!70}{%
    \parbox{\dimexpr\textwidth-2\fboxsep}{%
    \centering
        \textbf{Step 21}%
    }%
}
\textcolor{purple1}{\textbf{<think>} \\
It is now confirmed that the corresponding author of this paper is from the Second Affiliated Hospital of Chongqing Medical University, located in Southwest China (Chongqing). Next, I need to determine what honorary project this hospital received in 2025. Search result 1 mentions the ``Chongqing Integrated Traditional Chinese and Western Medicine `Flagship' Hospital Construction Project'', while search result 2 mentions the achievement of ``51 National Natural Science Foundation grants, a record high''. However, the question asks for an ``honorary project''; this could refer to the approval of the ``Chongqing Integrated Traditional Chinese and Western Medicine `Flagship' Hospital Construction Project'' or the achievement regarding National Natural Science Foundation grants?
Let's examine that news link (https://www.cqnews.net/1/detail/1331671774186954752/app/content\_1331671774186954752.html) first to check the specific content and the hospital's honors.\\
\textbf{</think>}}\\
\textcolor{black}{\textbf{</tool\_call>}}\\
\{``name'': ``visit'', ``arguments'': \{``url'': [``https://www.cqnews.net/1/detail/1331671774186954752/app/
content\_1331671774186954752.html''], ``goal'': ``Check the honorary projects obtained in 2025 by the hospital where the corresponding author of the study works''\}\} \\
\textcolor{black}{\textbf{</tool\_call>}}\\

\colorbox{gray!70}{%
    \parbox{\dimexpr\textwidth-2\fboxsep}{%
    \centering
        \textbf{Step 22}%
    }%
}
\textcolor{purple1}{\textbf{<think>} \\
The page visited just now did not mention any honorary projects obtained by the hospital in 2025. I need to search again for "Second Affiliated Hospital of Chongqing Medical University 2025 honors" or "Chongqing Medical University Second Affiliated Hospital 2025 received honors".\\
\textbf{</think>}}\\
\textcolor{black}{\textbf{<tool\_call>} \\
\{"name": "search", "arguments": \{"query": ["Second Affiliated Hospital of Chongqing Medical University 2025 honors"]\}\} \\
\textbf{</tool\_call>}}\\
\colorbox{gray!70}{%
    \parbox{\dimexpr\textwidth-2\fboxsep}{%
    \centering
        \textbf{Step 23}%
    }%
}
\textcolor{purple1}{\textbf{<think>} \\
Now I need to confirm which of these two items (51 National Natural Science Foundation projects, Chongqing Integrated Traditional Chinese and Western Medicine "Flagship" Hospital Construction Project) qualifies as the "honorary project". The question asks for the "honorary project obtained in 2025". The "Flagship Hospital" designation sounds more like an institutional honor or title, whereas the NSFC grants are research funding achievements. Let me search specifically for "Second Affiliated Hospital of Chongqing Medical University 2025 honorary project flagship hospital" to clarify.\\
\textbf{</think>}}\\
\textcolor{black}{\textbf{<tool\_call>} \\
\{"name": "search", "arguments": \{"query": ["Second Affiliated Hospital of Chongqing Medical University 2025 honorary project flagship hospital"]\}\} \\
\textbf{</tool\_call>}}\\
\textcolor{black}{\textbf{<tool\_response>}\\
Search Results are omitted...\\
\textbf{</tool\_response>}}\\

\colorbox{gray!70}{%
    \parbox{\dimexpr\textwidth-2\fboxsep}{%
    \centering
        \textbf{Step 24}%
    }%
}
\textcolor{purple1}{\textbf{<think>} \\
Let's review all the clues from the question again:
1. Phototherapy: Wavelength around 300 nm (Narrowband UVB or Excimer laser, 308/311 nm), used for Cutaneous T-cell Lymphoma (CTCL). DNA damage involves pyrimidine dimers (covalent cyclic structures between adjacent pyrimidine bases).
2. Repair System: Nucleotide Excision Repair (NER). The endonuclease component is XPF (ERCC4/FANCQ), which belongs to the Q complementation group of Fanconi Anemia (FA, a hereditary blood disorder).
3. Core substrate monoubiquitinated by the E3 ligase complex: FANCD2 (FA gene D2 type). Mutations here cause FA-D2 subtype, characterized by radial ray skeletal defects (radial bone absence/hypoplasia) in the upper limbs.
4. 2025 Research: Inhibition of the chromatin architecture protein DEK improves Fanconi Anemia. Corresponding authors include Professor Deng Jianchuan from the Second Affiliated Hospital of Chongqing Medical University (Southwest China).
5. Honorary project of this hospital in 2025: Search results mention the approval of the "Chongqing Integrated Traditional Chinese and Western Medicine 'Flagship' Hospital Construction Project" and a record-high 51 National Natural Science Foundation grants.
I need to definitively identify which one is referred to as the "honorary project". The "Flagship Hospital" status is typically an institutional honor/title awarded by the government, while NSFC grants are competitive research funding. I will search for "What is the 2025 honorary project of the Second Affiliated Hospital of Chongqing Medical University" to be sure.\\
\textbf{</think>}}\\
\textcolor{black}{\textbf{<tool\_call>} \\
\{"name": "search", "arguments": \{"query": ["What is the 2025 honorary project of the Second Affiliated Hospital of Chongqing Medical University"]\}\} \\
\textbf{</tool\_call>}}\\

\colorbox{gray!70}{%
    \parbox{\dimexpr\textwidth-2\fboxsep}{%
    \centering
        \textbf{Step 25}%
    }%
}
\textcolor{purple1}{\textbf{<think>} \\
Now I need to confirm the honorary project obtained by the Second Affiliated Hospital of Chongqing Medical University (the affiliation of the corresponding author of the DEK inhibition study) in 2025. Let me revisit the two previous links:
1. CQ News link regarding National Natural Science Foundation
2. Xinhua Net link regarding the Flagship Hospital
I will visit these two pages to determine which one represents the "honorary project".
\textbf{</think>}}\\
\textcolor{black}{\textbf{<tool\_call>} \\
\{"name": "search", "arguments": \{"query": ["What is the 2025 honorary project of the Second Affiliated Hospital of Chongqing Medical University"]\}\} \\
\textbf{</tool\_call>}}\\

\colorbox{gray!70}{%
    \parbox{\dimexpr\textwidth-2\fboxsep}{%
    \centering
        \textbf{Step 26}%
    }%
}
\textcolor{purple1}{\textbf{<think>} \\
Based on the information from the two pages:
1. The CQ News page states that in 2025, the hospital approved 51 National Natural Science Foundation projects, setting a historical high. This is a research achievement (project funding), not an honorary project title.
2. The Xinhua Net page states that the Second Affiliated Hospital of Chongqing Medical University was successfully selected for the "Chongqing Municipal Integrated Traditional Chinese and Western Medicine 'Flagship' Hospital Construction Project" list. This is the honorary project (title/construction project) obtained in 2025.
The question asks "What is the honorary project obtained in 2025?". Therefore, the "Chongqing Municipal Integrated Traditional Chinese and Western Medicine 'Flagship' Hospital Construction Project" is the honorary project (selection/approval as a flagship project), whereas the National Natural Science Foundation results are research project fundings.
Next, I need to confirm if the DEK inhibition study was indeed published in J Exp Med in 2025 by the team of Deng Jianchuan from the Second Affiliated Hospital of Chongqing Medical University. I will search for "Deng Jianchuan DEK Fanconi Anemia".
\textbf{</think>}}\\
\textcolor{black}{\textbf{<tool\_call>} \\
\{"name": "search", "arguments": \{"query": ["Deng Jianchuan DEK Fanconi Anemia"]\}\} \\
\textbf{</tool\_call>}}\\

\colorbox{gray!70}{%
    \parbox{\dimexpr\textwidth-2\fboxsep}{%
    \centering
        \textbf{Step 27}%
    }%
}
\textcolor{purple1}{\textbf{<think>} \\
Next, I will visit the NetEase article (JEM | Hou Yu's team reveals the mechanism of hematopoietic stem cell exhaustion in Fanconi anemia) to confirm the details about DEK inhibition and the corresponding authors.\\
\textbf{</think>}}\\
\textcolor{black}{\textbf{<tool\_call>} \\
\{"name": "visit", "arguments": \{"url": ["https://www.163.com/dy/article/JMIT15OQ0532BT7X.html"], "goal": "Check the corresponding authors, chromatin architecture protein, and hospital honors in this study"}\} \\
\textbf{</tool\_call>}\\

\colorbox{gray!70}{%
    \parbox{\dimexpr\textwidth-2\fboxsep}{%
    \centering
        \textbf{Step 28}%
    }%
}
\textcolor{purple1}{\textbf{<think>} \\
Now I have confirmed all the key information:\\
1. Phototherapy: Narrowband UVB (~311 nm) for Cutaneous T-cell Lymphoma (CTCL) causes cyclobutane pyrimidine dimers (covalent ring structures between adjacent pyrimidines), repaired by Nucleotide Excision Repair (NER).\\
2. Q Complementation Group: The NER endonuclease component XPF (also known as ERCC4 or FANCQ) belongs to the Q complementation group of Fanconi Anemia (FA).\\
3. E3 Ligase Substrate: The core substrate monoubiquitinated by the FA E3 ligase complex is FANCD2. Mutations in FANCD2 cause FA-D2 subtype, characterized by radial ray defects (e.g., absent or hypoplastic radius).\\
4. 2025 Study: A study published in \textit{Journal of Experimental Medicine} in 2025 found that inhibiting the chromatin architecture protein DEK restores hematopoietic stem cell function in Fanconi Anemia. The corresponding authors are from Chongqing Medical University, specifically the Second Affiliated Hospital (Southwest China).\\
5. Hospital Honor: The Second Affiliated Hospital of Chongqing Medical University received the honor of being selected for the "Chongqing Municipal Integrated Traditional Chinese and Western Medicine 'Flagship' Hospital Construction Project" in 2025.\\
\\
All clues align perfectly. I will now compile the final answer.\\
\textbf{</think>}}\\
\textcolor{black}{\textbf{<answer>}}\\
 Key Information Chain Summary:\\
1. **Phototherapy and DNA Damage Mechanism**: Narrowband UVB (approx. 311 nm) is used for phototherapy in Cutaneous T-cell Lymphoma (CTCL). The resulting DNA damage consists of cyclobutane pyrimidine dimers (covalent ring structures formed between adjacent pyrimidine bases), which are repaired by the Nucleotide Excision Repair (NER) system.\\
2. **Q Complementation Group and E3 Ligase Substrate**: The NER endonuclease component \textbf{XPF/ERCC4/FANCQ} belongs to the \textbf{Q complementation group} of the hereditary blood disorder \textbf{Fanconi Anemia (FA)}. The core substrate monoubiquitinated by the FA E3 ubiquitin ligase complex is \textbf{FANCD2}. Mutations in the gene encoding FANCD2 define the \textbf{FA-D2 subtype}, which is clinically characterized by upper limb developmental abnormalities, specifically radial ray defects such as absence or hypoplasia of the radius.\\
3. **2025 Research and Hospital Honor**: A 2025 study published in the \textit{Journal of Experimental Medicine} revealed that targeting the chromatin architecture protein \textbf{DEK} (e.g., using the nucleic acid aptamer DTA-64 to inhibit its function) can improve Fanconi Anemia. The corresponding authors of this study are affiliated with \textbf{Chongqing Medical University}, specifically the \textbf{Second Affiliated Hospital} (located in Southwest China). In 2025, the Second Affiliated Hospital of Chongqing Medical University received the honorary project designation of the \textbf{"Chongqing Municipal Integrated Traditional Chinese and Western Medicine 'Flagship' Hospital Construction Project"}, awarded jointly by the Chongqing Health Commission and the Administration of Traditional Chinese Medicine to enhance diagnostic and treatment capabilities in key areas like oncology through the integration of traditional and western medicine.\\
\textcolor{black}{\textbf{</answer>}}

\end{tcolorbox}
\end{document}